# Disentangling brain heterogeneity via semi-supervised deep-learning and MRI: dimensional representations of Alzheimer's Disease

**Authors:** Zhijian Yang[a,†], Ilya M. Nasrallah[b,†], Haochang Shou[c], Junhao Wen[a], Jimit Doshi[a], Mohamad Habes[d], Guray Erus[a], Ahmed Abdulkadir[a], Susan M. Resnick[e], David Wolk[f], Christos Davatzikos[a,*], &

[†] These authors contribute equally to this work

[a] *Center for Biomedical Image Computing and Analytics, University of Pennsylvania, Philadelphia, PA, USA*
[b] *Department of Radiology, University of Pennsylvania*
[c] *Department of Biostatistics, Epidemiology and Informatics, University of Pennsylvania*
[d] *Biggs Alzheimer's Institute, University of Texas San Antonio Health Science Center, USA*
[e] *Laboratory of Behavioral Neuroscience, National Institute on Aging*
[f] *Department of Neurology, University of Pennsylvania,*
*& for the iSTAGING consortium, for the Baltimore Longitudinal Study of Aging, for the Alzheimer's Disease Neuroimaging Initiative‡*

[*] Corresponding author: Christos Davatzikos, Christos.Davatzikos@pennmedicine.upenn.edu, 3700 Hamilton Walk, 7th Floor, Center for Biomedical Image Computing and Analytics, University of Pennsylvania, Philadelphia, PA 19104; https://www.med.upenn.edu/cbica/

[‡] Data used in preparation of this article were obtained from the Alzheimer's Disease Neuroimaging Initiative (ADNI) database (adni.loni.usc.edu). As such, the investigators within the ADNI contributed to the design and implementation of ADNI and/or provided data but did not participate in analysis or writing of this report. A complete listing of ADNI investigators can be found
at: http://adni.loni.usc.edu/wpcontent/uploads/how_to_apply/ADNI_Acknowledgement_List.pdf.

# ABSTRACT

Heterogeneity of brain diseases is a challenge for precision diagnosis/prognosis. We describe and validate Smile-GAN (**SeMI**-supervised c**L**ust**E**ring-Generative Adversarial Network), a novel semi-supervised deep-clustering method, which dissects neuroanatomical heterogeneity, enabling identification of disease subtypes via their imaging signatures relative to controls. When applied to MRIs (2 studies; 2,832 participants; 8,146 scans) including cognitively normal individuals and those with cognitive impairment and dementia, Smile-GAN identified 4 neurodegenerative patterns/axes: P1, normal anatomy and highest cognitive performance; P2, mild/diffuse atrophy and more prominent executive dysfunction; P3, focal medial temporal atrophy and relatively greater memory impairment; P4, advanced neurodegeneration. Further application to longitudinal data revealed two distinct progression pathways: $P1 \rightarrow P2 \rightarrow P4$ and $P1 \rightarrow P3 \rightarrow P4$. Baseline expression of these patterns predicted the pathway and rate of future neurodegeneration. Pattern expression offered better yet complementary performance in predicting clinical progression, compared to amyloid/tau. These deep-learning derived biomarkers offer promise for precision diagnostics and targeted clinical trial recruitment.

# 1. INTRODUCTION

Neurologic and neuropsychiatric diseases and disorders are often very heterogeneous in their neuroimaging and clinical phenotypes. Artificial intelligence methods, especially machine learning approaches, have shown great promise in deriving individualized neuroimaging signatures of anatomy, function and pathology that offer diagnostic and prognostic value[1]. However, until recently very little attention has been placed on using these methods to capture disease heterogeneity in interpretable ways, and on identifying distinct disease subtypes which might have different prognosis, progression patterns, and response to treatments. Toward this goal, a novel semi-supervised deep learning paradigm is presented herein (Fig. 1), referred to as Smile-GAN (**SeMI**-supervised c**L**ust**E**ring via **G**enerative **A**dversarial **N**etwork). Smile-GAN models disease effects via sparse transformations of normal anatomy, leveraging a GAN that is trained to synthesize realistic scans that is hard to be distinguished from real patient scans. Estimated latent variables capture neuroanatomical subtypes, which modulate this synthesis in an inverse consistent formulation.

Smile-GAN is used to identify the heterogeneity in cerebral neuroanatomy found across a spectrum from early cognitive impairment to dementia, using 8,146 scans from 2 different longitudinal studies used in the iSTAGING consortium[2]. (ADNI, the Alzheimer's Disease Neuroimaging Initiative, and BLSA, the Baltimore Longitudinal Study of Aging[3,4] both including 2 phases) obtained from 2,832 individuals. Alzheimer's disease (AD) is the most common neurodegenerative disease, affecting millions across the globe,[5] and accounts for the majority of cognitive decline in this study sample. The hallmark pathology of AD includes the presence of ß-amyloid neuritic plaques and tau protein-containing neurofibrillary tangles, which contribute to the characteristic neurodegeneration measured on magnetic resonance imaging (MRI). While diagnostic criteria have traditionally focused on the clinical syndrome, typically a predominately amnestic phenotype for AD and a pre-dementia phase called Mild Cognitive Impairment (MCI), recently there has been effort to define AD biologically based on the presence of amyloid deposition (A), tau deposition (T), and neurodegeneration (N), each characterized typically dichotomously as either absent (-) or present (+) and, thus, defining the AT(N) framework.[6] While useful, such binary implementations of this paradigm poorly capture the heterogeneity of cognitive impairment, such as known variability in AD topography or effects of common copathologies, including vascular disease and other comorbid neurodegenerative processes that might affect the 'N' dimension in distinct ways. This variability, along with patient resilience to neuropathology, plays an important role in the ultimate expression of cognitive decline in the individual and is therefore critical to understand when moving beyond group effects of disease to personalized diagnostics. Further, by more clearly identifying typical patterns and severity of neurodegeneration, including patterns more suggestive of underlying AD, such methods may allow improved selection of participants for clinical trials.

Several MRI biomarkers have been used to quantify neurodegeneration in AD. One of the most common is hippocampal volume,[7] a characteristic feature of typical AD. However, as other single region-of-interest (ROI) markers, it is neither specific, nor does it capture a complex atrophy pattern across the brain. Composite measures sensitive to the typical temporoparietal atrophy seen in AD, including various regional volumetric signatures[8] or machine learning metrics like SPARE-AD,[9,10] provide alternative measures of neurodegeneration that also capture relevant changes across multiple brain regions. These methods capture individual dimensions of neurodegeneration with high sensitivity, but do not provide information on alternative patterns when the primary pattern is poorly matched and may provide misleading information when applied to samples that do not match the target disease. Recently, novel data-driven methods to identify patterns of cerebral atrophy in AD and other neurodegenerative diseases have emerged, leveraging large neuroimaging data sets and novel machine learning methodology.

Clustering methods have been used to identify cross-sectional or temporal heterogeneity in patients. However, these approaches may identify clusters related to various disease-irrelevant confounding factors that influence inter-individual brain variations[11-13]. The semi-supervised method proposed here aims to overcome this limitation by effectively clustering differences between cognitively normal (CN) individuals and patients, thereby focusing on neuroanatomical heterogeneity of pathologic processes rather than heterogeneity that might be caused by a variety of confounding factors.[14,15] Deep learning, in particular, has made a notable leap in medical imaging applications.[16] Generative adversarial networks (GAN)[17] are well-known for learning and modeling complex distributions using a competition between two neural networks, and hence have been used to synthesize exceptionally realistic images. Herein they are used, to model disease effects.

Building on GAN-based models[18-21], we present a novel method (Smile-GAN), which captures different disease-related neuroanatomical patterns by generating realistic data by transforming neuroanatomical data of CN individuals. Via inverse-consistent latent variables, this synthesis is guided by disease-related neuroanatomical subtypes, which in turn are estimated from the data. Moreover, we extensively validate this method using simulated data as well as synthesized patterns of brain atrophy. In addition to leveraging the power of GANs to synthesize realistic data and hence to model disease effects, Smile-GAN does not make any assumptions of disease stages, which allows us to capture mixed pathologies via post-hoc analysis of expression of derived patterns and without any *a priori* assumptions. This is an important difference between Smile-GAN and alternative methods, e.g. SuStaIn model.[11]

We hypothesized that Smile-GAN would identify common patterns of neurodegeneration seen in patients along the AD pathway. We discovered 4 reproducible neuroanatomical patterns of atrophy across the spectrum of cognitive decline and developed ways to quantify the level of expression of each of these patterns in any individual, thereby arriving at a 4-dimensional coordinate system capturing key heterogeneity of the 'N' dimension in the AT(N) system. Critically, by measuring longitudinal trajectories within this coordinate system, we identified two distinct progression pathways which imply the presence of copathologies and heterogeneity of pathological processes. We identify baseline patterns with excellent predictive abilities for future neurodegenerative and clinical trajectories for individual participants.

## 2. RESULTS

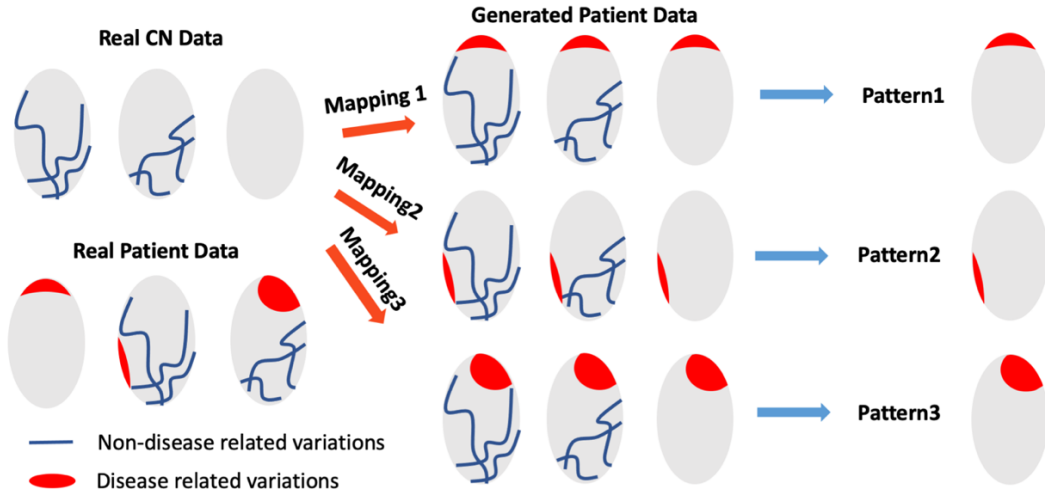

Fig. 1: Conceptual overview of Smile-GAN. Blue lines represent non-disease-related variations observed in both CN and patient groups. Red regions represent disease effects which only exist among patient groups. Smile-GAN finds neuroanatomical pattern types by means of clustering transformations from CN data to patient data.

## 2.1 Validation of Smile-GAN Model on Synthetic and Semi-synthetic Dataset

Experiments on a synthetic dataset (Supplementary Method 3.1) verified the ability of the model to capture heterogeneous disease-related variations while not being confounded by non-disease-related variation. Mapping functions captured all regions with simulated atrophy along each direction while almost perfectly avoiding all regions with much stronger simulated non-disease-related variations. (Supplementary Fig. 2(A)). Experiments on the semi-synthetic dataset, a real MRI dataset with synthesized brain atrophy, (Supplementary Method 3.2), further validated the ability of the model to avoid non-disease-related variability when they are more realistic. Moreover, the performance of the model was shown to be superior to other state-of-the-art semi-supervised clustering methods and traditional clustering methods in detecting simulated pattern types even with very small and variable atrophy rates. (See Supplementary Table 3)

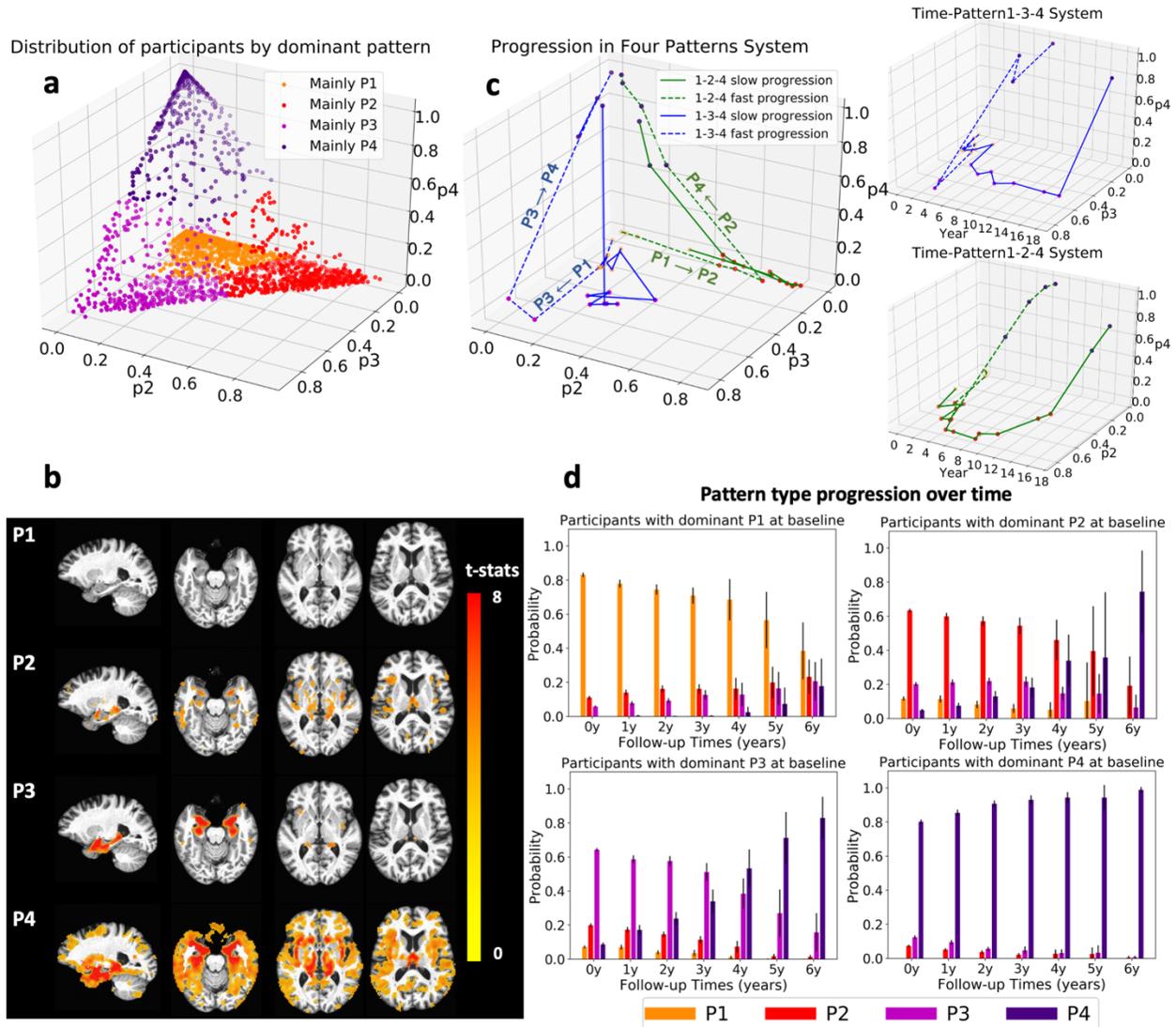

Fig. 2: Characterization of four atrophy patterns and two progression pathways of neurodegeneration. **a)** Visualization of participants' expression of four patterns in a three-dimensional coordinate. Pseudo-probabilities of belonging to each pattern reflect "levels of expression" (i.e., presence) of respective patterns and probabilistic subtype memberships. The origin point represents a pure P1 pattern, since the probabilities of belonging to different patterns sum up to 1, so probabilities of P2, P3 and P4 are sufficient for capturing expression of all 4 patterns. **b)** Voxel-wise statistical comparison between CN and participants predominantly belonging to the four patterns. False discovery rate (FDR) correction for multiple comparisons with p-value threshold of 0.05 was applied. **c)** Progression paths of four representative participants. One more dimension of time is added for visualization of progression speed. **d)** Bar plots of expression of the four patterns over time for each baseline pattern group. Mean values and 95 percent confidence interval of the mean calculated with t-distribution are reported for each year.

## 2.2 Four Patterns and Two Progression Pathways

**Four Patterns.** Smile-GAN identified four patterns of brain atrophy in cognitively impaired participants, using the CN as reference group. The four pattern types were found to be reproducible using the hold-out cross validation experiment. Figure 2a graphs probabilities in a 3-dimensional space, with each participant colored based on the dominant pattern (a 3D system is sufficient for visualization of these 4 patterns, because the respective probabilities sum up to 1). Participants within each pattern share similar atrophy patterns, which are shown by voxel-based group comparison results between the CN group and the group of participants dominated by each pattern (Fig. 2B). From these results, we can visually interpret the four imaging patterns as: i) P1, preserved brain volume, exhibits no significant atrophy across the brain compared to CN; ii) P2, mild diffuse atrophy, with widespread mild cortical atrophy with not particularly pronounced medial temporal lobe atrophy; iii) P3, focal medial temporal lobe atrophy, showing localized atrophy in the hippocampus and the anterior-medial temporal cortex with relative sparing elsewhere; iv) P4, advanced atrophy, displaying severe atrophy over the whole brain including severe temporal lobe atrophy. These four patterns were reproduced when training the model with only participants with positive ß-amyloid (Abeta) status (Supplementary Fig. 3), indicating that these four pattern probabilities capture common variation observed in participants who show evidence of AD-related neuropathological change.

**Two Progression Pathways** Figure 2d reveals evolution of pattern probabilities over time. Participants with P1 features at baseline may express increasing probability of P2 or P3 in the short term followed by later expression of the P4 pattern. Participants with dominant P2 or P3 expression at baseline show variable minor expression of the other pattern (other-pattern probability range 0 to 0.5). Both P2 and P3 have increasing P4 probability at later time points, but do not develop significant expression of the other P3 or P2 pattern, respectively. Participants who initially had highest values in P4 only show stronger expression of P4 over time. From these results, we conclude that P1-2-4 and P1-3-4 are two general MRI-progression pathways of neurodegeneration. Figure 2c displays detailed progression paths of some representative participants in the pattern-dimension system. These examples demonstrate that despite following similar progression pathways, participants may have difference in pattern purity and progression speed.

### 2.3 Neuropathological and Clinical Characteristics of Pattern Types

**Amyloid/Tau/Pattern/Diagnosis.** Most of CN participants had negative Abeta status (A-) and express P1 (Fig. 3a). P1 also included the largest number of cognitively impaired but nondemented participants, classified in BLSA/ADNI as MCI, with disproportionately amyloid negative status compared to the other three patterns. There was a comparable amount of MCI/Dementia participants with P2 and P3 (144 and 178) and they had similar distributions in amyloid status, predominately amyloid positive (66.9% and 72.1%). P4 participants were mostly amyloid positive (84.0%) and were rarely CN (3.3%). Pattern membership can be used to classify participants based on the AT(N) criteria, providing insight into the stage of the disease, resilience, and presence of copathology. Those placed along the AD continuum are further subgrouped into early neurodegeneration (P3), advanced (P4) neurodegeneration, or a P2 group of atypical atrophy that may be classified as N- or N+ by other methods. In Fig. 3b, participants are grouped as normal, as falling along the typical AD continuum, as AD with dominant copathology or as suspected non-AD pathology (SNAP) based on patterns and Abeta/phospho-tau (pTau) status. A+T+ participants tend to have more severe neurodegeneration than A+T- participants, as expected. SNAP is suggested by the presence of neurodegeneration and/or tau deposition without evidence of amyloid deposition.

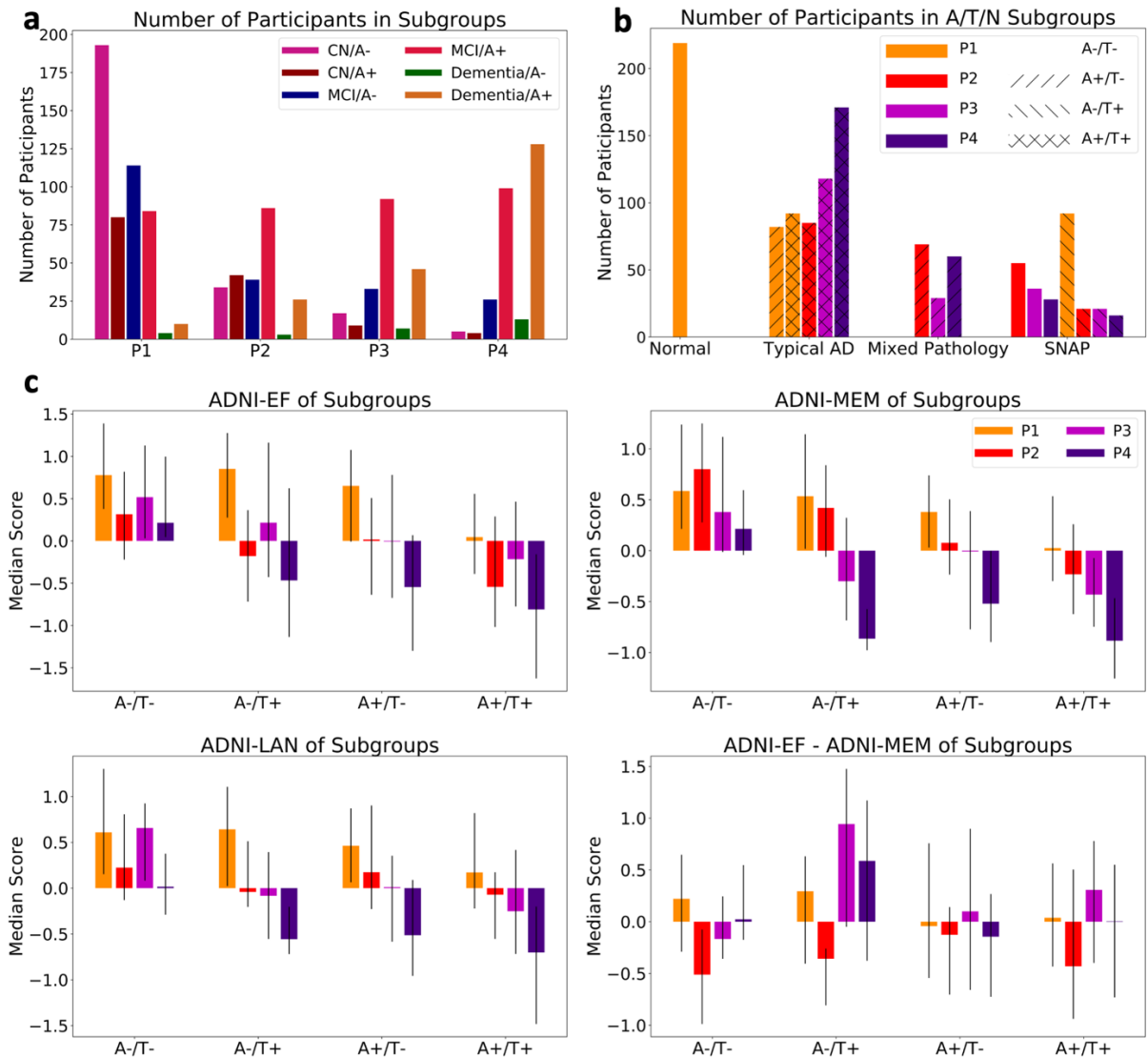

Fig. 3: Participants grouping and cognitive performance of subgroups. **a)** Number of participants grouped by diagnosis, amyloid status and pattern. **b)** AT(N) categorization based on participants' patterns and CSF Abeta/pTau status. Based on patterns, N is classified as normal (P1), not typical of AD (P2), or characteristic of AD (P3/P4). **c)** Cognitive performance of participants by pattern. Median scores with bars showing the first and the third quartile. (A: Abeta; T: pTau)

**MRI and Clinical Characteristics** Statistical comparisons of MRI and clinical characteristics were conducted among A+ cognitively impaired participants with different dominant patterns. (Supplementary Table 5). P4 and P1 participants showed significantly higher and lower WML volume, respectively, (50.6 mm and 34.1 mm, p<0.001), but there was no significant difference between P2 and P3 (46.3 mm and 45.1 mm, p=0.863). P3 and P4 participants showed significantly lower hippocampal volumes (0.54 and 0.52, respectively for P3 and P4, versus 0.61 for P1, p<0.001 for both comparison). Certain features were the highest in P3 participants: ApoE ε4 allele carrier rate (78%) and tTau (341.1) and pTau (34.9) levels. Differences in cognitive scores were revealed by comparing across participants with the same A/T status (Fig. 3C and Supplemental Table 5). Regardless of A/T status, P1 participants had much better performance across cognitive domains while P4 participants had the poorest performance. P2 participants showed worse performance in executive function than P3 participants but had better function in memory. This distinction was more clearly seen examining the difference between ADNI-EF and ADNI-MEM, which was significantly different between P2 and P3 participants (p=0.039 for A-T-, p=0.003 for A-/T+ and p<0.001 for A+/T+). Investigation of special subgroups showed additional features of disease. Within A+P3 participants, CN and impaired groups had similar tTau (p=0.33) and pTau (p=0.48) levels, suggesting comparable AD pathologic change. However, A+P3 CN participants had significantly longer education (p=0.001), higher hippocampal volumes (p=0.018), and less expression of the P3 pattern probabilities (p=0.048) compared to A+P3 impaired participants, suggesting that higher cognitive reserve and less neurodegeneration may account for the preservation of cognitive function in this relatively small group (n=19, Supplemental Table 6). Among T-P1 participants with MCI/Dementia, amyloid status was not associated with significant differences in clinical measures shown in Supplemental Table 7, showing that amyloid pathology alone was not resulting in greater impairment in the A+ group. On the other hand, A+T+P1 participants did show significantly worse cognitive performance (Figure 2C) along with greater atrophy in hippocampus (p=0.011) and significantly lower P1-probability (p<0.001), suggesting early neurodegeneration that might be related to underlying AD pathology.

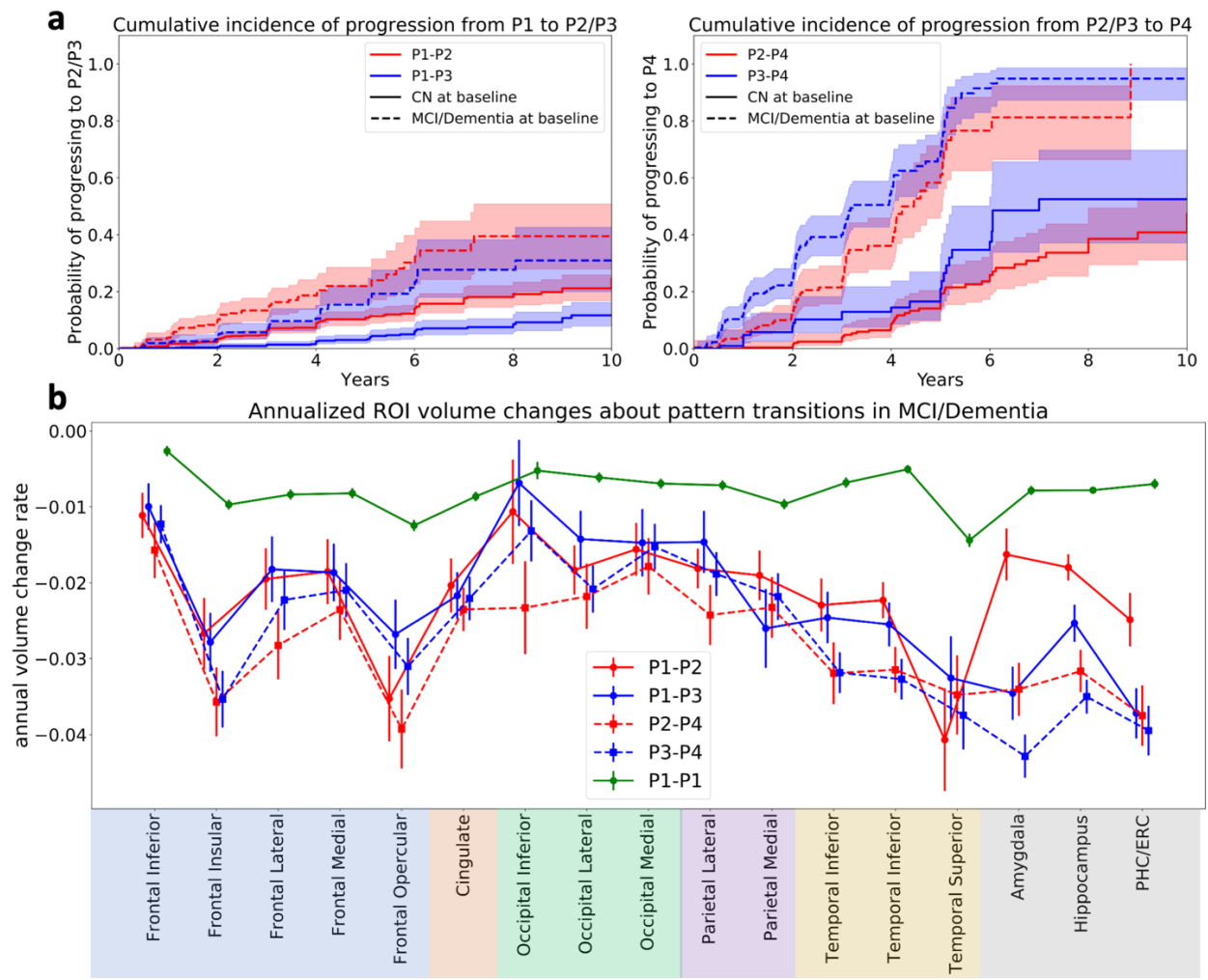

Fig. 4: Analysis of longitudinal pattern progression. **a)** Cumulative incidence of pattern progression. The line styles indicate the diagnosis at baseline. 95% confidence intervals are shown with cumulative incidence curves **b)** Annual atrophy rate in selected regions along different paths. Data within three years before pattern change or last follow-up point (for 1-1 participants) were utilized and random intercept model with time as fixed effect was used to derive annual volume change rate with respect to baseline volume. Standard error is shown with the corresponding ratio. (PHC: Parahippocampal gyrus; ERC: Entorhinal cortex)

## 2.4 Longitudinal Progression of Pattern Types

Cumulative incidence curves in Fig. 4a show that progression from P1 at baseline is faster to P2 than to P3, and progression from P3 at baseline is faster to P4 than progression from P2 at baseline. These relationships hold regardless of cognitive diagnosis at baseline, although baseline CN have much slower rate of progression than those with baseline cognitive impairment. Figure 4b displays differences in volume changes of selected regions among distinct progression pathways. First, participants who persist in P1 show much lower longitudinal atrophy in all selected regions. Participants progressing from P1 to P3 show faster medial temporal lobe atrophy while those progressing from P1 to P2 show faster frontal and occipital atrophy. There is an acceleration in medial temporal lobe atrophy associated with the P2-P4 transition. While classified together as P4 pattern, distinct regional atrophy can be observed between P4 participants who progressed from P2 versus P3, reminiscent of those earlier patterns (Supplementary Fig. 3).

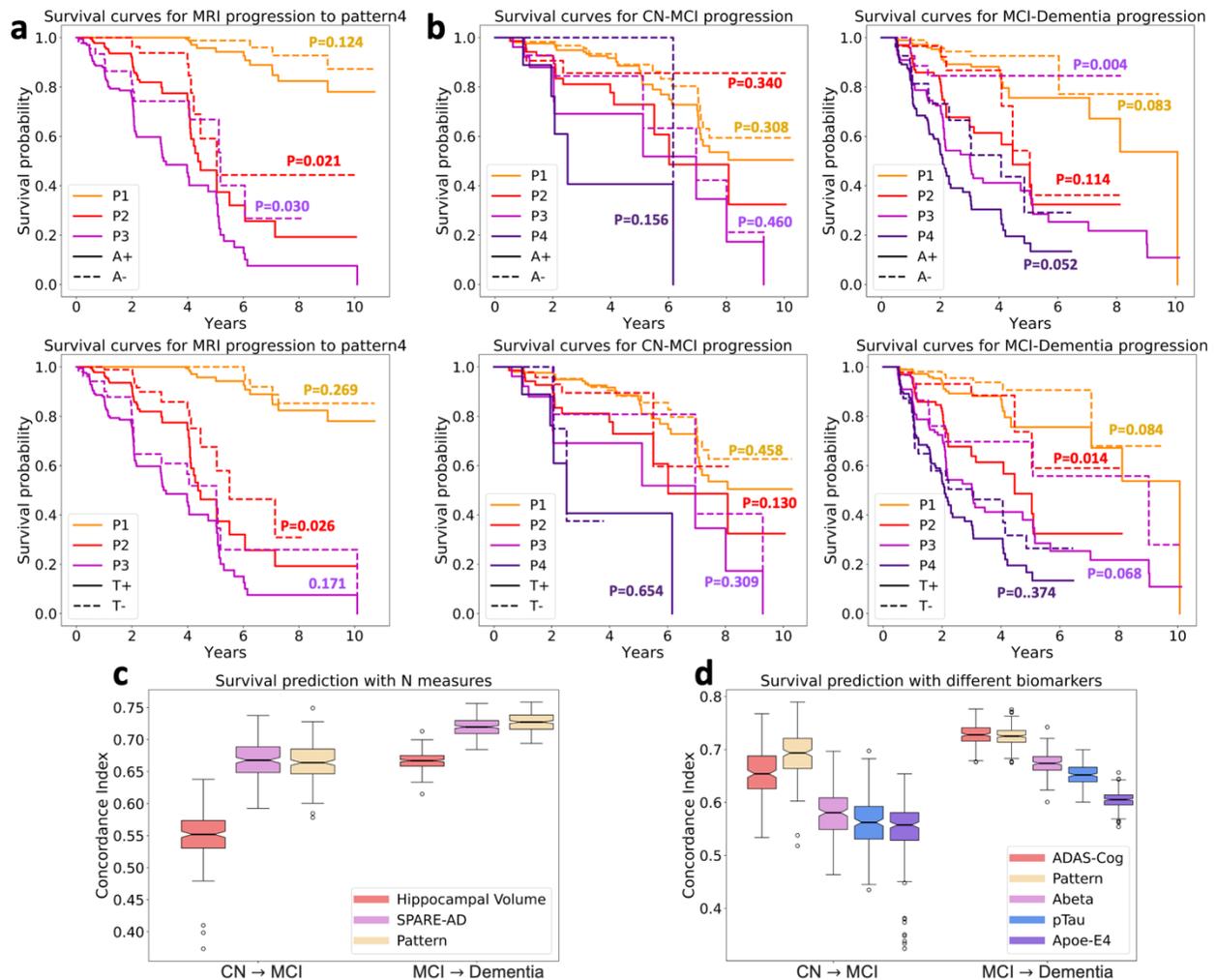

Fig. 5: Predictive ability of patterns. **a)** Survival curves for neurodegeneration progression to P4; **b)** Survival curves for clinical diagnosis progression from CN to MCI and from MCI to Dementia. For both (a) and (b), survival curves are stratified by both initial dominated patterns and Abeta/pTau status; p values indicate statistical significance of difference between positive and negative Abeta/pTau status; **c,d)** Concordance Index (C-Index) measures the performance of Cox-proportional-hazard model in predicting clinical conversion time (from CN to MCI and MCI to Dementia). Different biomarkers are utilized as features of the model for evaluation of their predictive powers.

## 2.5 Predictive Power of Pattern Probabilities

**Prediction of MRI progression.** Survival curves in Fig. 5a illustrate that participants' baseline pattern expression are associated with the risk of the conversion to p4. Abeta and pTau status at baseline further differentiate higher versus lower risk of future conversion to P4. Moreover, among baseline P1 participants, baseline P2 and P3 probabilities predict longitudinal progression pathways and progression speed. Using Cox-proportional-hazard models, the baseline probabilities of P2 were able to discriminate participants with different event time of progressing from P1 to P2 and achieve an average concordance index (C-Index) of 0.823±0.022 on the validation set. Similar analyses using baseline P3 probabilities to predict risk of progressing to P3 achieved an average C-index of 0.844±0.024. Thus, baseline P2 and P3 probabilities were able to indicate the chance and direction of P1 participants progressing to P2 or P3 within two to five years (see Supplementary Table 4). Prediction performance worsened after the fifth year and the optimal threshold for predicting progression along either pathway decreased with time (see Supplementary Table 4).

**Prediction of Clinical Progression (Change in Diagnosis).** Clinical categorizations of CN, MCI, and dementia provide useful information on functional status. Survival curves in Fig. 5b reveal that, even with similar Abeta or pTau status at baseline, participants with different pattern types show different progression rates for clinical categories. The discrepancy is greater in the MCI to Dementia progression than for the CN to MCI progression, which occurs less frequently across groups. However, only for participants with P2 and P3 at baseline, pTau and Abeta status add significant discrimination power to the survival rate from MCI to Dementia. Furthermore, pattern probabilities at baseline have comparable predictive power with the SPARE-AD score,[9] a previously validated predictive biomarker of AD neurodegeneration, while both outperformed hippocampal volume (see Fig. 5c). Also, compared with other biomarkers including APOE genotype, ADAS-cog score, Abeta and pTau measures, pattern probabilities show either comparable or superior performance in prediction of both CN to MCI and MCI to Dementia progression (Fig. 5d).

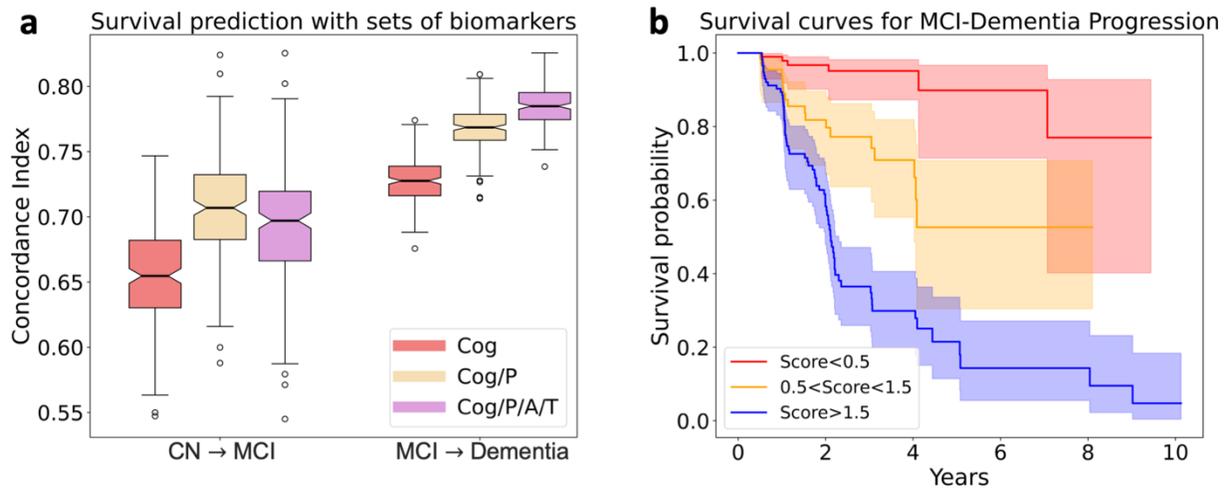

Fig. 6: Prediction of clinical diagnosis progression with composite biomarkers. **a)** Biomarkers were added successively into features set based an order of accessibility. Concordance Index (CI) measures the performance of Cox-proportional-hazard model in predicting clinical conversion time (from CN to MCI and MCI to Dementia). Different sets of biomarkers are utilized as features of the model for evaluation of their predictive powers. **b)** Survival curves stratified by composite scores (A, T, Pattern, ADAS-Cog jointly predicting outcome in cross-validated fashion) for one randomly split validation set. 95% confidence intervals are shown with survival curves. (A: Abeta; T: pTau, P: Pattern, Cog: ADAS-Cog score)

**Composite Score for Risk of Clinical Progression.** With ADAS-Cog score, the most easily ascertained biomarker, as the only feature, the Cox-proportional-hazard model was able to achieve an average C-Index of $0.654\pm0.034$ for prediction of CN to MCI progression and $0.728\pm0.020$ for prediction of MCI to Dementia progression through cross-validation. Further addition of pattern probabilities derived from T1 MRI significantly boosted average C-Indices for both tasks to $0.702\pm0.042$ and $0.768\pm0.017$ respectively. However, inclusion of Abeta/pTau derived from relatively expensive PET scans did not bring significant additional improvement to prediction performance (Fig. 6a). With all these biomarkers utilized together, we could construct a composite score indicating risk of clinical progression that was able to predict survival time from MCI to Dementia with an average C-index of $0.785\pm0.016$, on randomly split validation sets. Examples of survival curves stratified by the composite score for one randomly split validation set are shown in Fig. 6b.

## 3. DISCUSSION

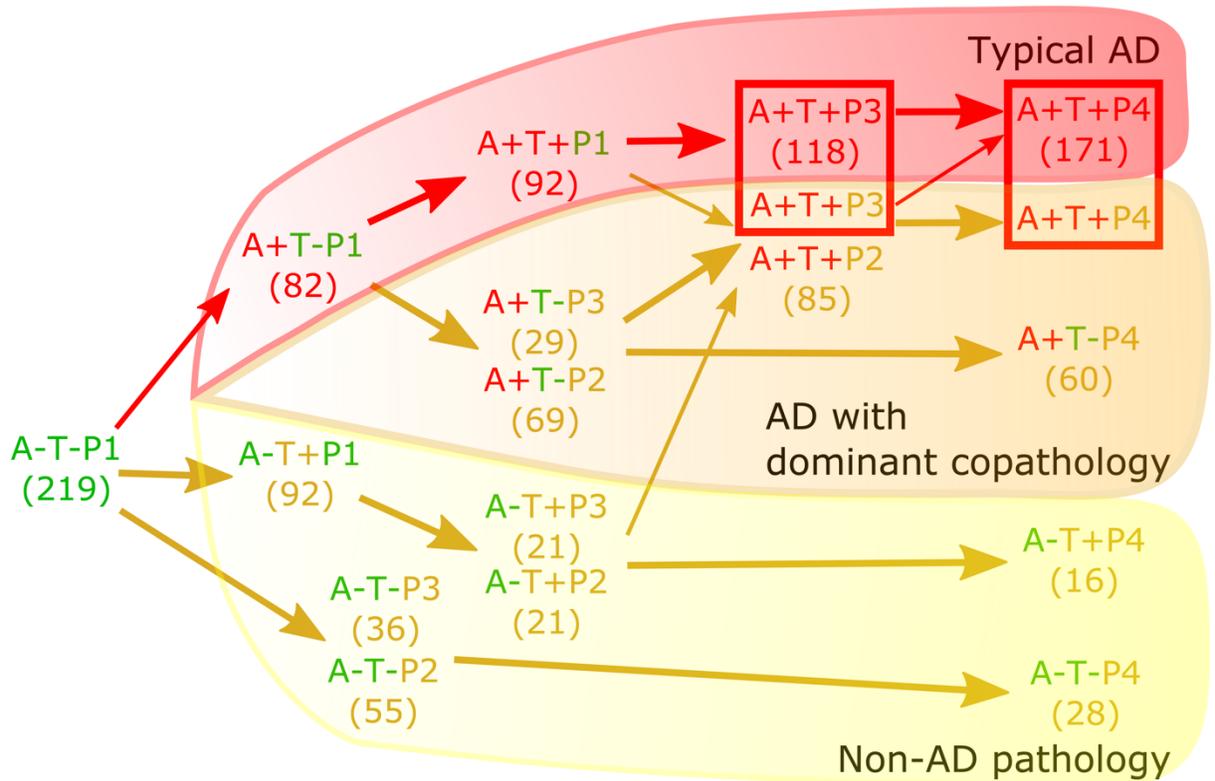

Fig. 7. Implications for ATN framework. Cascade of biomarkers can follow a 'canonical AD' pathway, which is the most represented in the ADNI sample. The interrelationships of patterns with amyloid/tau status identifies another large group with the presence of AD pathology and significant or even dominant copathology as well as groups with suspected non-AD pathology. These pathways also indicate that certain 'typical' AD neurodegenerative phenotypes may in some cases be driven by copathology. For example, boxed phenotypes are typical for AD, however there are several potential paths whereby copathology may be the dominant cause of the neurodegenerative pattern. Red arrows and biomarkers indicate typical AD features. Yellow arrows and biomarkers indicate pathology that is atypical for AD. Green features indicate the normal condition. Numbers in parentheses indicate number of ADNI participants in each category at baseline.

We have developed a new deep learning approach, Smile-GAN, which disentangles anatomical heterogeneity and defines subtypes of neurodegeneration by learning to generate mappings from images of cognitively normal individuals to images of patients. Compared with unsupervised methods,[11-13] Smile-GAN has a significant advantage in avoiding non-disease related confounding variations, thereby identifying neuroanatomical patterns associated with the pathologic processes. This stems from the fundamental property of Smile-GAN to cluster the transformations from normal to pathologic anatomy, rather that clustering patient data directly. Also, the deep learning-based Smile-GAN can easily handle high dimensional ROI data. Thus, no pre-processing ROI selection is required, and the model is able to fully capture variations in all subdivided ROIs. Moreover, in contrast with other semi-supervised methods,[14,15] Smile-GAN makes no assumption about data distribution and data transformation linearity, and in validation experiments was found to be more robust to mild, sparse, or overlapping patterns of pathology (neurodegeneration, herein). Critically, pattern probabilities given by Smile-GAN are easily interpretable continuous biomarkers reflecting the neuroanatomical expression of respective patterns. These advantages of Smile-GAN allow versatile characterization of pattern types related to both severity and heterogeneity of pathological effects.

An important characteristic of Smile-GAN is that it keeps pattern and stage separate in its formulation. In particular, Smile-GAN simply measures the degree of expression of a given pattern and doesn't assume that a person who belongs to a subtype continues along that subtype with increasing stage measures (e.g., as in the SuStaIn method[11]). Staging can be inferred at a second-level analysis of the degree of expression of each of the 4 patterns, or a linear or nonlinear combination of them. This allows for the possibility of an individual expressing multiple subtypes at the same time, reflected by substantial magnitude of respective probabilities, as well as of defining complex and nonlinear relationships between pattern-based stage and clinical outcomes of interest, which can vary depending on the outcome (e.g., from various cognitive or clinical measures to staging estimates that inform clinical trial recruitment).

Application of Smile-GAN to MRI data from a sample enriched with AD pathology identified 4 patterns of brain atrophy expressed in participants across the AD spectrum. These patterns range from mild to advanced atrophy and define two progression pathways. One pathway, here termed the P1-3-4 pathway, shows early atrophy in the medial temporal lobe that is typical for AD. The second pathway, P1-2-4, shows early diffuse mild cortical atrophy with MTL sparing that is a less typical pattern for AD. The end stage for both pathways is an advanced atrophy stage (P4). Pattern membership is associated with differences in cognitive test performance, with P2 having relatively more executive dysfunction, P3 showing greater memory impairment, and P4 showing the worst performance across domains. These patterns also have implications for speed and direction of progression, with early pattern features predictive of the future pattern of neurodegeneration and pattern features predictive of progression from CN to MCI and MCI to dementia. Critically, pattern expression was the most important predictor of clinical progression, showing comparable or stronger predictive ability than other N measures and biomarkers (Fig. 5). Synergistically, pattern expression, A, T and ADAS-Cog provided outstanding cross-validated prediction of clinical progression on an individual basis (Fig. 6b), underlining the potential significance of this combined predictive index for patient management, for clinical trial recruitment, and for evaluation of treatment response.

While the patterns are relatively distinct in regional specificity and severity, the underlying pathophysiology is more complex. P1 indicates that no significant neurodegeneration is present, from any etiology. Yet a significant number of participants with P1 pattern still had cognitive impairment as marked by their MCI stats, with even a few with dementia, both with and without evidence of amyloid and tau deposition. These participants likely have reduced cognitive reserve and possibly non-neurodegenerative contributions to MCI/dementia. The P2 group shows mild diffuse atrophy and is likely a group inclusive of multiple mild or early pathologies, such as atypical/cortical presentations of AD, other early neurodegenerative processes or mild diffuse atrophy related to chronic systemic disease. P2 is not disproportionately enriched for vascular disease, a common comorbidity for primary neurodegenerative diseases, at least as measured by WML volumes which were relatively similar across P2-P3-P4. Regardless of etiology, expression of a P2 pattern is akin to concepts of advanced brain aging or decreased brain reserve. While P3 is predominately early typical AD within the enriched ADNI sample, this also likely includes other pathologies such as limbic-associated TDP-43 encephalopathy (LATE). P4 appears to be a composite of advanced or 'end-stage' neurodegeneration patterns. While fully typical of advanced AD, this pattern is also seen in participants with cognitive decline without amyloid or tau deposition, indicating a late-stage similarity of brain atrophy across multiple pathologies.

With the growing utilization of the AT(N) framework, these patterns provide a means to quantify neurodegeneration into a few informative categories rather than as a binary measure. As such, it allows categorization of neurodegeneration as absent (P1), atypical for AD (P2), typical for AD (P3) or advanced (P4). Together with A/T status, the dynamics of pattern expression shows both severity of disease and identifies reasonably distinct and reasonably sized subgroups with differing balance of AD and non-AD pathology (Fig. 7). While the patterns have significant prognostic implications in isolation, interpretation in the setting of amyloid, tau, and cognitive status illuminates several important features of AD and related disorders, specifically groups that may have impaired brain reserve (P2) or those who show resilience and normal cognition in the setting of AD-related neurodegeneration (A+T+P3/4). These groups could be used to enrich for typical AD pathology for clinical trials, reduce the need for ascertaining certain biomarkers, and identify interesting subgroups for focused evaluation, such as for genetic factors of resilience. For example, to recruit a group with early AD neurodegeneration, one could initially select those with P3 pattern on MRI and ascertain A/T biomarkers only in this group.

The Smile-GAN pattern approach has several advantages. It captures biologically relevant atrophy patterns that are few in number, providing meaningful detail on neurodegeneration without adding significant complexity. The Smile-GAN method is a data-driven approach that can be applied on features extracted from data beyond neuroimages, clustering patients effectively based on disease related feature changes from normal group to patient group. Therefore, it is generalizable to any diseases and disorders that have reproducible patterns of changes in imaging or other biomedical data, including but not limited to other neurodegenerative and neuropsychiatric diseases[22].While there are modest time and computational requirements for training the model, the training process is only performed once, and subsequent calculation of individual pattern scores using an existing model is rapid. There are limitations to the method and our implementation. First, pathologies or stages of disease than have common phenotypic expression may not be distinctly separated (Supplemental Figure 3), while rare or subtle patterns of atrophy may not be distinctly learned by the model. It is possible that larger and more diverse training data may allow identification of more pattern types. The performance of the four-pattern model in this study was derived and evaluated using data from the ADNI and BLSA studies, which have high and low prevalence of AD, respectively, and relatively low prevalence of non-AD neurodegeneration. Direct application of this model to a memory-center population with mixed neurodegenerative disease has not been evaluated.

Patterns identified using semi-supervised clustering with generalized adversarial networks provide useful information about the severity and distribution of neurodegeneration across the AD spectrum. Baseline patterns are predictive of the future pattern of neurodegeneration as well as clinical progression to MCI and dementia. These patterns could augment research and clinical assessments of participants and patients with cognitive decline and contribute to a dimensional characterization of brain diseases and disorders.

# 4. METHODS

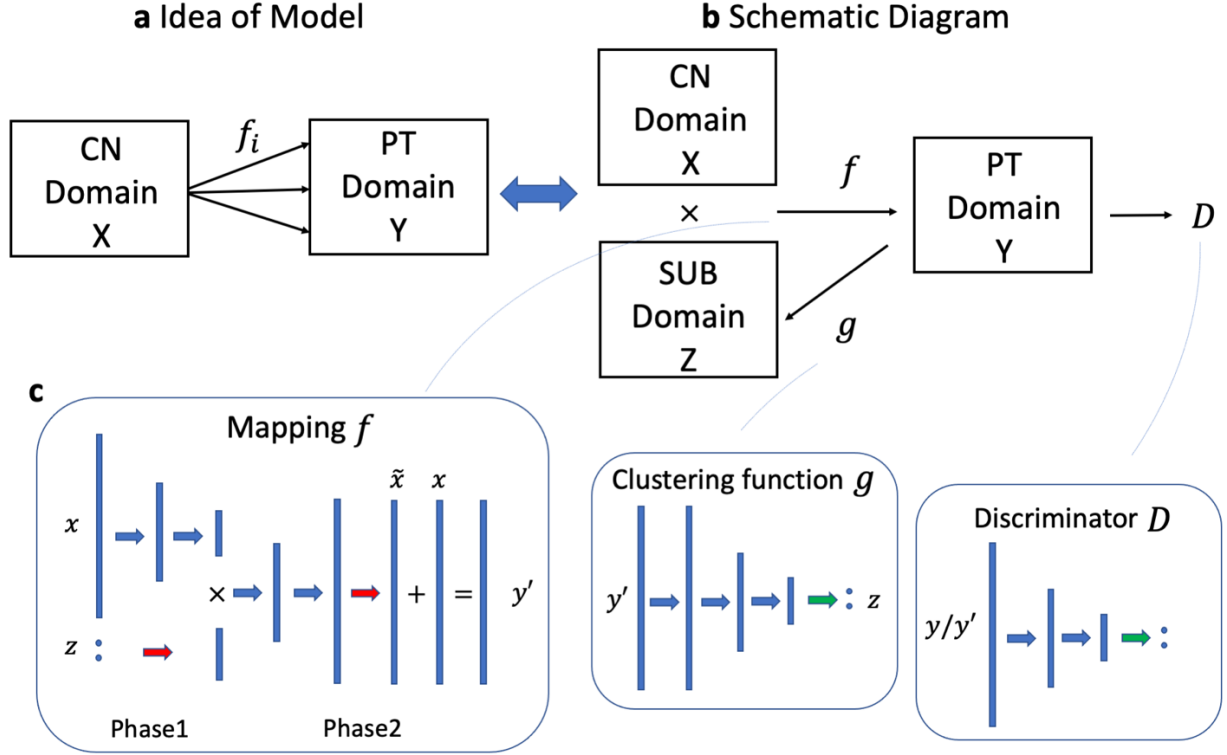

Fig. 8: Schematic diagram and network architectures. **a)** General idea behind Smile-GAN. The model aims to learn several mappings from the CN domain to the PT domain **b)** Schematic diagram of Smile-GAN. The idea of the model is realized by learning one mapping from joint domain $X \times Z$ to $Y$, while learning another function $g: Y \to Z$. CN: cognitive normal control, PT: patient, Sub: pattern subtype. **c)** Network architecture of three functions: blue arrow represents one linear transformation followed by one leaky rectified linear unit function, green arrow represents one linear transformation followed by one softmax function, red arrow represents only one linear transformation.

## 4.1 Smile-GAN Model

Smile-GAN is a novel Generative Adversarial Network (GAN) architecture for clustering a group (in our case patients) based on their multi-variate differences (in our case regional volumes derived from MRI) to a reference group (in our case healthy controls). The general structure of Smile-GAN is shown in Fig. 8. To sum up, the primary concept of the model is to learn one-to-many mappings from the CN domain $X$ to the patient (PT) domain $Y$. The idea is equivalent to learning one mapping function, $f: X \times Z \to Y$, which generates synthesized PT data $y' = f(x, z)$ from the joint domain $X \times Z$, while enforcing the indistinguishability between PT data and synthesized PT data. Put simply, given one same value for subtype variable, $z$, the mapping $f(\cdot)$ generates image data that match data of patients of similar subtype mix. Here, $Z = \{\alpha: \sum_{i=1}^{M} \alpha_i = 1\}$, referred as the subtype (SUB) domain, is a class of vectors with dimension $M$ ($M = 4$ was found to be optimal in our experiments). We denote the distribution of the aforementioned variables as $x \sim p_{CN}, y \sim p_{PT}, y' \sim p_f, z \sim p_{Sub}$ respectively. The variable $z$, independent from $x$, takes values from a subclass of domain $Z$ and can be encoded as a one-hot vector with value 1 being placed at any position with equal probability (i.e., $1/M$). In addition to the mapping function, an adversarial discriminator $D$ is introduced to distinguish between real PT data $y$ and synthesized PT data $y'$, thereby ensuring that the mappings $f$ generate image data that are indistinguishable from real patient data.

The fact that a number of functions can potentially achieve equality in distributions makes it hard to guarantee that the mappings learned by the model are closely related to the underlying pathology progression. Moreover, during the training procedure, the mapping function backboned by the neural network tends to trivially ignore the Sub variable $z$. Therefore, with the assumption that there is one true underlying function for real PT variable $y = h(x, z)$, Smile-GAN aims to boost the mapping function $f$ to be approximate to the true underlying function $h$, by constraining the function class via three types of regularization: 1) we encourage sparse transformations, 2) enforce Lipschitz continuity of functions, 3) introduce another function $g: Y \rightarrow Z$ to the model structure. The latter is a critical part of the algorithm's ability to cluster the data, as it requires that the mapping functions identify sufficiently distinct imaging patterns in the Y domain, which would allow the inverse mapping $g(\cdot)$ to estimate the correct subtype in the domain. More details about regularization terms and clustering inference of function $g$ are stated in Supplementary Method 1.

The objective of Smile-GAN is a combination of adversarial loss[17] and regularization terms. First, the adversarial loss[17] aims at matching the distribution synthesized PT data, $p_f$, to the distribution of real PT data, $p_{PT}$, which can be denoted as:

$$L_{GAN}(D, f) = E_{y \sim p_{PT}}[\log(D(y))] + E_{z \sim p_{Sub}, x \sim p_{CN}}[1 - \log(D(f(x, z)))]$$
$$= E_{y \sim p_{PT}}[\log(D(y))] + E_{y' \sim p_f}[1 - \log(D(y'))]$$

where the mapping $f$ attempts to transform CN to synthetically generated PT data so that they follow similar distributions as real PT data. The discriminator $D$, providing a probability that $y$ comes from the real data rather than the generator, is trying to identify the synthesized PT data and distinguish it from the real PT data. Therefore, the discriminator attempts to maximize the adversarial loss function while the mapping $f$ attempts to minimize against it. The corresponding training process can be denoted as:

$$\min_f \max_D L_{GAN}(D, f) = E_{y \sim p_{PT}}[\log(D(y))] + E_{y' \sim p_f}[1 - \log(D(y'))]$$

Second, the regularization terms include the change loss and cluster loss, both serving to constrain the function space where $f$ is learned from. The change loss is defined as:

$$L_{change}(f) = E_{x \sim p_{CN}, z \sim p_{Sub}}[\|f(x, z) - x\|_1]$$

By denoting $l_c$ to be the cross-entropy loss with $l_c(a, b) = -\sum_{i=1}^{k} a^i \log b^i$, we define the cluster loss as:

$$L_{cluster}(f, g) = E_{x \sim p_{CN}, z \sim p_{Sub}}[l_c(z, g(f(x, z)))]$$

With the aforementioned losses, we can write the full objective as:

$$L(D, f, g) = L_{GAN}(D, f) + \mu L_{change}(f) + \lambda L_{cluster}(f, g)$$

where $\mu$ and $\lambda$ are two hyperparameters that control the relative importance of each loss function during the training process. Through this objective, we aim to find the mapping function $f$ and clustering function $g$ such that:

$$f, g = \arg\min_{f,g} \max_{D} L(D, f, g)$$

More implementation details of the model, including network architecture, training details, algorithm, and training stopping criteria are presented in Supplementary method 2.

### 4.2 Study and Participants

The Alzheimer's Disease Neuroimaging Initiative (ADNI, http://www.adni-info.org/) study is a public-private collaborative longitudinal cohort study which has recruited participants categorized as cognitively normal, MCI, and AD participants through 4 phases (ADNI1, ADNIGO, ADNI2).[23] ADNI has acquired longitudinal MRI, cerebrospinal fluid (CSF) biomarkers, and cognitive testing. The Baltimore Longitudinal Study of Aging, neuroimaging substudy, Imaging substudy, has been following participants who are cognitively normal at enrollment with imaging and cognitive exams, since 1993. A total number of 1718 ADNI participants (819 ADNI1 and 899 ADNI-GO/ADNI2) and 1114 BLSA participants were included in the study. Details of both studies including number classified as CN/MCI/Dementia at baseline, number of participants with CSF Abeta/Tau biomarkers, length of follow up, age, gender, APOE genotype are included in Table4.

| Study | CN | MCI | Dementia | Mean Follow-up (years) | Gender (% of male) | Age | APOE E4 Carriers | CSF Abeta/Tau Available |
|---|---|---|---|---|---|---|---|---|
| ADNI1 | 229 | 397 | 193 | 2.2 (1.7-3.1) | 58.2% | 75 (71-80) | 48.8% | 415 |
| ADNI2/GO | 297 | 452 | 150 | 2.1 (1.1-4.0) | 53.3% | 74 (68-79) | 43.6% | 824 |
| BLSA | 1094 | 11 | 9 | 4.0 (0.0-6.0) | 47% | 67 (58-76) | 25% | 0 |

Table 1. Details of ADNI and BLSA studies. For age and length of follow-ups, median value with first and third quartile are reported. APOE E4 carriers include heterozygotes and homozygotes.

### 4.3 MRI Data Acquisition and Processing

1.5 T and 3T MRI data were acquired from both ADNI and BLSA study introduced above. A fully automated pipeline was applied for processing T1 structural MRIs. T1-weighted scan of each participant is first corrected for intensity inhomogeneities.[24] A multi-atlas skull stripping algorithm was applied for the removal of extra-cranial material.[25] For the ADNI study, 145 anatomical regions of interest (ROIs) were identified using a multi-atlas label fusion method[26]. For the BLSA study, this method was combined with harmonized acquisition-specific atlases[27] to derive 145 ROIs. Phase-level cross-sectional harmonization was applied on regional volumes of 145 ROIs to remove site effects.[28] For visualization of disease patterns, tissue density maps, referred as RAVENS (regional analysis of volumes examined in normalized space[29]) were computed as follows. Individual images were first registered to a single subject brain template and segmented into grey matter (GM) and white matter (WM) tissues. RAVENS maps encode, locally and separately for each tissue type, the volumetric changes observed during the registration.

### 4.4 Data Separation and Preparation

After preprocessing, baseline ROI data of 297 CN and 602 cognitively impaired participants from ADNI2/GO participants were selected as the discovery set for training and validation of the model. longitudinal ROI data from follow-up visits of all participants from ADNI and BLSA were used for further clinical analysis, including both participants whose baseline data were used for model training and those who were completely independent of the discovery set. For analysis requiring measures of CSF Abeta/pTau, only ADNI participants with these two biomarkers were included. Otherwise, all participants from ADNI and BLSA study were incorporated for analysis.

Before being used as features for the Smile-GAN model, ROI volumes were residualized and variance-normalized. To correct age and sex effects while keeping disease-associated neuroanatomical variations, we estimated ROIs-specific age and sex associations among 297 CN participants using a linear regression model. All cross-sectional and longitudinal data were then residualized by age and sex effects. Then, all ROI volumes were further normalized with respect to 297 CN participants in the discovery set to ensure a mean of 1 and standard deviation of 0.1 among CN participants for each ROI.

### 4.5 Cognitive, Clinical, CSF Biomarker and Genetic Data

We used additional clinical, biofluid, and genetic variables, including CSF biomarkers of amyloid and tau, APOE genotype, and cognitive test scores, provided by ADNI. These measures were downloaded from the LONI website. A total of 1194 participants from ADNI have CSF measurements of ß-amyloid, total tau, and phospho-tau cutoffs for amyloid status based on ß-amyloid measures and for tau status based upon phospho-tau measures were previously defined[30] and used to categorize participants as positive or negative for cerebral amyloid and tau deposition. Tau measures are also presented as continuous variables. Composite cognitive scores across several domains have been previously validated in the ADNI cohort, including a memory composite (ADNI-MEM),[31] an executive function composite (ADNI-EF),[32] and a language composite (ADNI-LAN).[33]

White matter lesion (WML) volumes were calculated from both ADNI and BLSA using inhomogeneity-corrected and co-registered FLAIR and T1-weighted images and a deep learning-based segmentation method[34] built upon the U-Net architecture,[35] with the convolutional layers in the network replaced by an Inception ResNet architecture.[36] The model was trained using a separate training set with human-validated segmentation of WML. WML volumes were first cubic rooted. Then phase-level cross-sectional harmonization was applied on them to reduce site effects.

### 4.6 Pattern Memberships and Probabilities Assignments.

Smile-GAN model assigns $M$ probability values to each participant, with each probability corresponding to one pattern type and the sum of $M$ probabilities being 1. Based on the $M$ probability values, we can further assign each participant to the dominant pattern type, determined by the maximum probability. The "optimal" $M$ was chosen during a cross-validation (CV) procedure based on the clustering reproducibility or stability. Specifically, we ran 20 repetitions of repeated holdout CV for $M=3$ to 5, with 10% of the data being left out. Of note, $M=2$ generally stratified the data into mild and severe atrophy patterns, which is not clinically interesting. For each repetition, we randomly left out 10% of the discovery set to add variability. We used the

Adjusted Rand Index (ARI)[37] to quantify the clustering stability of the 20 repetitions/models. The highest mean pair-wise ARI, $0.437 \pm 0.054$, was reached at $M=4$, which lead to the results presented herein. After choosing the "optimal" $M$, we reran Smile-GAN 30 times with all available data in the discovery set and the trained models will be used for external validation and analysis. In order to find the best correspondence among cluster assignments across the 30 experiments, we calculated the mean pair-wise ARI values for each resultant model. The one with the highest ARI was chosen as the template and the pattern types learned by all other models were reordered so that their clustering results achieved the highest overlap with that of the template. After reordering, the average probability of each pattern across all 30 models was taken as the probability of the corresponding pattern for each participant. We then applied these learned models to longitudinal data of all CN/MCI/Dementia participants and obtained probabilities of four patterns for all visits of each participant.

### 4.7 Statistical Analysis

To visualize the brain signatures of four patterns, we utilized all cross-sectional data of MCI/Dementia participants in the discovery set and performed voxel-wise group comparisons (i.e., CN vs each pattern) via AFNI 3dttest[38] using voxel-wise tissue density (RAVENs) maps[29]. To access longitudinal progression trajectories of pattern assignment, we grouped for each of the four patterns those participants with probability larger than 0.5. We then compared how the pattern probabilities change over time for each of the four groups by calculating pattern probability for P1-P4 of all within group who have data available in a given time interval (i.e., X year - X+1 year). Those who had more than one data point in the selected time interval only contributed once through mean probabilities of all those visits. The demographic variables, APOE genotype, CSF biomarker levels, cognitive test scores, WML volumes and pattern probabilities were compared both across pattern types and within pattern types. Only participants from the ADNI study whose Abeta/pTau status was available at baseline were included for comparison. For categorical variables, the chi-squared test was used to identify differences between subgroups. For other quantitative variables, a one-way ANOVA analysis was performed for group comparison.

To assess the risk of converting from P1 into P2 or P3, we conducted time-to-event survival analysis to evaluate the risk pattern conversions. In particular, we treated P2 and P3 as competing events and used Aalen-Johansen estimator to generate cumulative incidence curves for P1 to P2/P3 progression. For all other cumulative incidence curves and survival curves corresponding to pattern progression and diagnosis transformation, we applied a nonparametric Kaplan-Meier estimator and used the log rank test to compare difference in survival distributions between groups.

For all survival analysis, participants were assigned into one pattern at baseline or labeled as progressing to one pattern only if the corresponding pattern probability is greater than 0.5. A few participants not reaching this threshold in any pattern at baseline were discarded to avoid noise in the analysis.

### 4.8 Evaluation of Patterns' Predictive Ability

We further conducted analyses to evaluate the predictive ability of baseline pattern probabilities in the prediction of future pattern changes. Also, we compared them with other measures of neurodegeneration (N measures) and clinical biomarkers in prediction of diagnosis transitions.

For pattern progression prediction, we selected all 940 participants who had longitudinal follow-ups and P1>0.7 at baseline to avoid trivial prediction tasks. First, the Cox-proportional-hazard model with baseline P2 or P3 probability as the only feature was utilized to predict survival curves from P1 to P2 or P1 to P3 respectively. We ran the two-fold cross validation 100 times and derived the concordance-index on validation sets. Second, to predict risk of pattern progression and progression pathways of P1 participants at specific time points T, we directly used P2 probability and P3 probability at baseline as an indication of risk without further fitting any additional models. For each time T from 2 years to 8 years, we generated a binary indicator with 0 representing not progressing to P2 till T and 1 representing who have already progressed to P2 before T and directly used baseline P2 probabilities to discriminate these two groups. The exact same process was also done for P3. Area under the receiver operator characteristic curve (AUC) values were calculated for both P2 and P3 at different time T. Optimal discrimination thresholds, at which true positive rate (TP) plus false positive rate (FP)=1, were reported for two different progression pathways.

To predict clinical diagnosis changes, we selected out 1178 CN participants and 921 participants categorized as MCI at baseline who had longitudinal follow-ups. First, to compare Patterns with other N measures, we again utilized the Cox-proportional-hazard model with different N measures as features to predict CN-MCI and MCI-Dementia survival curves. Two-fold cross validation was run 100 times to derive the concordance-index on validation sets. Then, to compare the prognostic powers of Pattern, Abeta, pTau, APOE genotype and ADAS-Cog, we reduced samples to 380 CN and 568 MCI participants who had these biomarkers. Each biomarker was used independently as the only feature for training the model.

For all prediction tasks in this section, baseline pattern assignments and progression labelling followed the same rule introduced in section 4.7 if not specifically annotated.

### 4.9 Biomarker selection and Composite Score Construction

Finally, we evaluated predictive powers of different combinations of biomarkers mentioned above. Following the order of accessibility, ADAS-Cog, pattern probabilities derived from T1 MRI, Abeta/pTau derived from PET scan were added successively to the feature set for training the Cox-proportional-hazard model and the same experimental procedure were implemented as introduced in section 4.8. A composite score indicating the risk of clinical progression can be derived with all biomarkers introduced above. Using Pattern-probabilities/Abeta/pTau/ADAS scores at baseline as features, the trained Cox-proportional-hazard model was applied to the validation set to derive the partial hazard as the composite score.

## 5. Acknowledgement

The iSTAGING consortium is a multi-institutional effort funded by NIA by RF1 AG054409. The Baltimore longitudinal study of aging has been funded by HHSN271201600059C, Intramural Research Program of the National Institutes of Health (NIH). ADNI is funded by the National Institute on Aging, the National Institute of Biomedical Imaging and Bioengineering, and through generous contributions from the following: AbbVie, Alzheimer's Association; Alzheimer's Drug Discovery Foundation; Araclon Biotech; BioClinica, Inc.; Biogen; Bristol-Myers Squibb Company; CereSpir, Inc.; Cogstate; Eisai Inc.; Elan Pharmaceuticals, Inc.; Eli Lilly and Company; EuroImmun; F. Hoffmann-La Roche Ltd and its affiliated company Genentech, Inc.; Fujirebio; GE Healthcare; IXICO Ltd.; Janssen Alzheimer Immunotherapy

Research & Development, LLC.; Johnson & Johnson Pharmaceutical Research & Development LLC.; Lumosity; Lundbeck; Merck & Co., Inc.; Meso Scale Diagnostics, LLC.; NeuroRx Research; Neurotrack Technologies; Novartis Pharmaceuticals Corporation; Pfizer Inc.; Piramal Imaging; Servier; Takeda Pharmaceutical Company; and Transition Therapeutics. The Canadian Institutes of Health Research is providing funds to support ADNI clinical sites in Canada. Private sector contributions are facilitated by the Foundation for the National Institutes of Health (www.fnih.org). The grantee organization is the Northern California Institute for Research and Education, and the study is coordinated by the Alzheimer's Therapeutic Research Institute at the University of Southern California. ADNI data are disseminated by the Laboratory for Neuro Imaging at the University of Southern California.

## 6. Data Availability

The data that support the findings of this study are available from their respective institutions, but restrictions apply to the availability of these data, which were used under license for the current study, and so are not publicly available. Data may however be available from the authors upon reasonable request and with permission.

# 1. SUPPLEMENTARY METHOD

## 1.1 Smile-GAN Regularization.

Regularizations mentioned in this section serve to constrain the function class where the mapping function $f$ is sampled from, so that it is truly meaningful while matching the distribution. The change loss is to control the distance of transformations. We assume that only some specific regions will be affected as disease progresses along each direction, which means that the true underlying transformation only changes some regions while keeping the rest unchanged. To encourage sparsity, we define the change loss to be the $l_1$ distance between the synthesized PT data and the original CN data:

$$L_{change}(f) = E_{x \sim p_{CN}, z \sim p_{Sub}}[\|f(x,z) - x\|_1]$$

The rest regularizations are based on Lipschitz continuity of the mapping function $f$ and clustering function $g$. First, with function $f$ being $K$-Lipschitz continuous, we have that, for fixed Sub variable $z = a$ and $\forall x_1, x_2 \in X$, $\|f(x_1, a) - f(x_2, a)\|_2 \leq K\|x_1 - x_2\|_2$. Thus, by controlling the constant $K$, the same mapping direction will preserve original distances among CN data by transforming them into a compact cluster but not scattering them into the PT domain. Moreover, with function $g$ being $K$-Lipschitz continuous, we derive that, $\forall z_1, z_2 \sim p_{Sub}, z_1 \neq z_2$ and $\bar{x} \sim p_{CN}$, $\|f(\bar{x}, z_1) - f(\bar{x}, z_2)\|_2$ is lower-bounded by $\frac{\sqrt{2}}{K} - \frac{1}{K}(\|g(f(\bar{x}, z_1)) - z_1\|_2 + \|g(f(\bar{x}, z_2)) - z_2\|_2)$:

$$\begin{aligned}
\|f(\bar{x}, z_1) - f(\bar{x}, z_2)\|_2 &\geq \frac{1}{K}\left(\|g(f(\bar{x}, z_1)) - g(f(\bar{x}, z_2))\|_2\right) \\
&\geq \frac{1}{K}\left(\|z_1 - z_2\|_2 - \|g(f(\bar{x}, z_1)) - z_1\|_2 - \|g(f(\bar{x}, z_2)) - z_2\|_2\right) \\
&= \frac{\sqrt{2}}{K} - \frac{1}{K}(\|g(f(\bar{x}, z_1)) - z_1\|_2 + \|g(f(\bar{x}, z_2)) - z_2\|_2)
\end{aligned}$$

Therefore, we can control differences among mapping directions to be non-trivial (i.e., same CN data is mapped to significantly different PT data along distinct directions) by minimizing the distance between sampled Sub variable $z$ and reconstructed Sub variable $g(f(x,z))$. We, thus, define another cluster loss to be the cross-entropy between sampled $z$ and reconstructed $g(f(x,z))$. By denoting $l(a,b) = -\sum_{i=1}^{k} a^i \log b^i$, we can write the cluster loss as:

$$L_{cluster}(f,g) = E_{x \sim p_{CN}, z \sim p_{Sub}}[l(z, g(f(x,z)))]$$

Function $g$ here is considered as the approximation of posterior distribution $P(z|f(x,z))$. Therefore, minimization of cross entropy-loss can be also interpreted as maximizing the mutual information between synthesized PT variable $y'$ and Sub variable $z$ as shown in Remark 1. In this sense, the mapping function is forced to best utilize information of Sub variable while also keeping mutual information between transformed data and original CN data by controlling transformation distance.

Moreover, with all constraints imposed on function $f$ and the assumption $p_{PT} = p_f$, we consider that $f$ satisfies all necessary conditions and is a good approximation of the underlying function $h$ such that $f(x,z) \approx h(x, \sigma(z))$ for some $\sigma \in \Omega$, where $\Omega$ is the class of all permutation functions which changes the order of $M$ elements in vector $z$. By reordering the results given by models, we simply ignore the permutation function and write as $f(x,z) \approx h(x,z)$, without loss of generality. For any new PT data $y_i = h(x_i, z_i) \sim p_{PT}$ coming in, we can have $g(y_i) = g(h(x_i, z_i)) \approx g(f(x_i, z_i))$, whose results were trained to be close to $z_i$. Therefore, function $g$ can be used as a clustering function on unseen PT data.

**Remark 1.** By minimizing the cluster loss $L_{cluster}$ defined above, we are maximizing a lower bound of the mutual information between Sub variable $z$ and synthesized PT data $y' = f(x,z)$. Considering $Q(z|y') = g(y')$ to be an approximation of distribution $P(z|f(x,z))$, mutual information denoted by $I$ and entropy denoted by $H$, we can derive that:

$$\begin{aligned}
I(z; f(x,z)) &= H(z) - H(z|f(x,z)) \\
&= E_{y' \sim f(x,z)} \left[ E_{z' \sim P(z|y')} [\log P(z'|y')] \right] + H(z) \\
&= E_{y' \sim f(x,z)} \left[ D_{KL}(P(\cdot|y') || Q(\cdot|y')) + E_{z' \sim P(z|y')} [\log Q(z'|y')] \right] + H(z) \\
&\geq E_{y' \sim f(x,z)} \left[ E_{z' \sim P(z|y')} [\log Q(z'|y')] \right] \\
&= E_{z \sim p_{AT}, y' \sim f(x,z)} [\log Q(z'|y')] \\
&= E_{z \sim p_{AT}, x \sim p_{CN}} [\log Q(z'|f(x,z))] \\
&\approx \frac{1}{n} \sum_{i=1}^{n} \langle z_i, \log g(f(x_i, z_i)) \rangle
\end{aligned}$$

The fifth and sixth line follows the Lemma 5.1 in Info-GAN[1] and law of the unconscious statistician (Lotus) Theorem respectively. Therefore, we have the mutual information bounded below by $\frac{1}{n}\sum_{i=1}^{n} \langle z_i, \log g(f(x_i, z_i)) \rangle$. Maximization of this lower bound is equivalent to minimizing the cluster loss $L_{cluster}$.

### 1.2 Smile-GAN Implementation Details

#### 1.2.1 Network Architecture

To improve the rate of convergence, the mapping function, instead of directly transforming the CN data to the synthesized PT data, first learns a change in the CN data and then takes the sum of them to obtain the synthesized PT data. Therefore, the architecture of the mapping function $f$ can be divided into two phases as shown in Main Fig. 7C. In the first phase, the CN data and the Sub variable are mapped to latent representations with the same dimension through encoder and decoder, respectively. The second phase has one decoding structure mapping the dot-product of two representations to the change $\tilde{x}$, which is added to the CN data $x$ to generate the synthesized PT data. The discriminator $D$ and the clustering function $g$ have similar encoding structures, with $D$ mapping PT/synthesized PT data to prediction vector with dimension 2 while the encoder $g$ mapping the synthesized PT data to a Sub representation. More details are shown in Table1 and Table2.

Table1: Architecture of mapping function $f$

|  | Layer | Input Size | Bias Term | Leaky Relu $\alpha$ | Output Size |
|---|---|---|---|---|---|
| Phase1 (Encoder) | Linear1+Leaky-Relu | 145*1 | No | 0.2 | 72*1 |
|  | Linear2+Leaky-Relu | 72*1 | No | 0.2 | 36*1 |
| Phase1 (Decoder) | Linear1+Sigmoid | M*1 | Yes | NA | 36*1 |

|            | Layer               | Input Size | Bias Term | Leaky Relu $\alpha$ | Output Size |
|------------|---------------------|------------|-----------|---------------------|-------------|
|            | Linear1+Leaky-Relu  | 36*1       | No        | 0.2                 | 72*1        |
| Phase2     | Linear2+Leaky-Relu  | 72*1       | No        | 0.2                 | 145*1       |
|            | Linear3             | 145*1      | No        | NA                  | 145*1       |

Table 2: Architecture of discriminator $D$ and clustering function $g$

|               | Layer               | Input Size | Bias Term | Leaky Relu $\alpha$ | Output Size |
|---------------|---------------------|------------|-----------|---------------------|-------------|
|               | Linear1+Leaky-Relu  | 145*1      | Yes       | 0.2                 | 72*1        |
| Discriminator | Linear2+Leaky-Relu  | 72*1       | Yes       | 0.2                 | 36*1        |
|               | Linear3+Softmax     | 36*1       | Yes       | NA                  | 2*1         |
|               | Linear1+Leaky-Relu  | 145*1      | Yes       | 0.2                 | 145*1       |
|               | Linear2+Leaky-Relu  | 145*1      | Yes       | 0.2                 | 72*1        |
| Clustering    | Linear3+Leaky-Relu  | 72*1       | Yes       | 0.2                 | 36*1        |
|               | Linear4+Softmax     | 36*1       | Yes       | NA                  | $M$*1       |

### 1.2.2 Training Details

We ensure Lipschitz continuity of functions $f$ and $g$ by performing weight clipping[2]. With $\Theta$ representing the space where weights of function $\theta_f$ and $\theta_g$ lie in, the compactness of $\Theta$ implies the K-Lipschitz continuity of functions $f$ and $g$, where K only depends on $\Theta$. The compactness of $\Theta$ is achieved by clapping the weights to a fixed box ($\Theta = [-c, c]^d$). In the implementation, c is empirically chosen to be 0.5. Further relaxation of the bound does not make much difference to results.

We set two parameters to be $\mu = 5$ and $\lambda = 9$ for all experiments. Also, we performed gradient clip for each iteration to avoid the explosion of gradient during the training process. For optimization, we used ADAM optimizer[3] with learning rate 0.0004 for Discriminator $D$ and 0.002 for mapping $f$ and clustering function $g$. $\beta_1$ and $\beta_2$ are 0.5 and 0.999, respectively.

### 1.2.3 Algorithm

Detailed training procedure of Smile-GAN is disclosed by Algorithm 1.

**Algorithm 1:** Smile-GAN training procedure. $l_c$ represents cross entropy loss and $e_i$ represents a one hot vector with 1 at $i_{th}$ component.
**while** *not meeting stopping criteria or reaching max_epoch* **do**
    **for** *all batches* $\{x_i\}_{i=1}^m, \{y_i\}_{i=1}^m$ **do**
        *Sample m integers* $\{a_i\}_{i=1}^m$ *with* $a_i \sim$ *discrete-U (1, M) and let* $z_i = e_{a_i}$
        **Update weights of discriminator** $D$**:** Use ADAM to update $\theta_D$ with gradient:
        $\nabla_{\theta_D} \frac{1}{m} \sum_{i=1}^m [(l_c(D(y_i), e_1) + l_c(D(f(x_i, z_i), e_0)))]$
        **Update weights of mapping function** $f$**:** Use ADAM to update $\theta_f$ with gradient:
        $\nabla_{\theta_f} \frac{1}{m} \sum_{i=1}^m [(l_c(D(f(x_i, z_i), e_1) + \lambda l_c(g(f(x_i, z_i), z_i) + \mu \| f(x_i, z_i) - x_i \|_1)]$
        **Update weights of clustering function** $g$**:** Use ADAM to update $\theta_g$ with gradient:

$$\nabla_{\theta_g} \frac{1}{m} \sum_{i=1}^{m} [(g(f(x_i, z_i), z_i)]$$
$$(\theta_f, \theta_g) = clip((\theta_f, \theta_g), -c, c)$$
  **end**

**end**

### 1.2.4 Stopping Criteria

For the real application, since the ground truth of patterns is unknown, we adopt an approximation of the Wasserstein distance (WD) as one metric for monitoring the training process and choosing the stopping point. For Smile-GAN, instead of deriving WD from optimization, we used the closed-form formula to compute the distance. For stopping criteria, we assume that, in the CN domain and in all subpopulations of the PT domain, the lower-dimensional representation of each data point (ROIs) is sampled from a multivariate Gaussian distribution. Though this assumption might be strong, it does enable us to estimate the WD quickly.

To be more specific, for each of $M$ Sub variables, $z = z_i$ and for all samples in CN domain $\bar{X} = \{x_1, x_2, \cdots, x_n\}$, we calculate the mean vector $m_1^i$ and covariance matrix $C_1^i$ of $f(\bar{X}, z_i)$. Also, from samples in PT domain Y, we take out the subset $Y_i = \{y_j\}^i$ such that $g(y_j)$ has highest value at position $i$ for all $y_j \in Y^i$. For this subset, we calculate the mean vector $m_2^i$ and covariance matrix $C_2^i$. With mean vectors and covariance matrices, we can compute the 2nd Wasserstein distance using the formula for two multivariate gaussian measure:

$$W_2(\mu_{f(X,z_i)}, \mu_{Y_i}) = \|m_1^i - m_2^i\|_2^2 + \text{trace}(C_1^i + C_2^i - 2\left(C_2^{i\frac{1}{2}} C_1^i C_2^{i\frac{1}{2}}\right)^{\frac{1}{2}})$$

If we further assume that all features are independent, we can derive diagonal covariance matrices which make the computation even faster. Based on our experiments on synthetic and semi-synthetic datasets, these assumptions do not affect monitoring the training process.

Moreover, to deal with cases when inconsistencies exist between WD and model performance, we also derive two other metrics: alteration quantity (AQ), which represents the number of participants whose dominated pattern type alter in the last five epochs. A small AQ represents high stability of the model. Lastly, the cluster loss, indicating the performance of clustering function $g$, is also considered as part of the stopping criteria.

### 1.3 Validation of Smile-GAN model

### 1.3.1 Synthetic Test

Simulated data were generated in a low dimensional space (i.e., 145 ROIs). For each participant, the 145 ROIs were simulated by sampling from a normal distribution N (1, 0.1). In total, 1200 participants were generated independently and then randomly split into two half-split sets, with each (600) being CN and pseudo-PT group, respectively. The atrophy simulation was only introduced for pseudo-PT participants, which were further divided into 3 pattern types with the same number of participants (200). For each pattern type, the values of specific pre-selected ROIs were decreased by 20%. Moreover, to simulate confounding non-disease-related effects, we randomly sampled 200 participants from both CN and pseudo-PT respectively and decreased the values by 40% in some other ROIs which make confounding factors much stronger than true patterns. The simulation ground truth for the confounding patterns and the 3 pattern types are shown in Fig. 2(A) (i) and (ii), respectively. Note that overlapping of ROIs across patterns was imposed to better follow the nature of atrophy.

To add variability of the clustering performance, we repeated the simulated data generation and ran the experiment independently 20 times. We first checked the clustering accuracy and validated the potential of WD for monitoring the raining process. Then we investigated the ROIs captured by mapping functions $f$ along different directions. We calculated the mean difference between CN and synthesized PT generated along $K$ mapping directions and inspected ROIs with a significant decreasing in values.

### 1.3.2 Synthesized Brain Atrophy Experiments

We included 526 CN from baseline T1 MRI in ADNI 1 and 2 databases. CN participants are split into two sets, 200 as the CN group, 326 as the pseudo-PT group. 326 pseudo-PT participants are further divided into three sets. Atrophy in medial temporal lobe was introduced to the first set. Atrophy in selected global regions was introduced to the second set. For the third set, a combination of these two patterns was introduced. Different atrophy strength level was tested by decreasing the chosen ROIs' value by 30%,10-30% and 10-20%, respectively. Semi-simulated data can incorporate more realistic non-disease-related variations among participants for validation.

We first validated the potential of WD for monitoring the training process on semi-synthetic data whose distribution is closer to the real dataset while may violating the assumption we made for stopping criteria. Second, we compared the performance of Smile-GAN with two other semi-supervised methods, HYDRA[4] and CHIMERA[5], and also two basic clustering methods, K-means and GMM on these same tasks through the holdout validation. Each time running the model, we randomly split 80% data as the training set to add variability. For one holdout analysis, we repeat this procedure C times and the final clustering membership was determined by a consensus clustering strategy across models trained on C different splits. Because of differences in required training time, C is chosen to be 5 for the Smile-GAN model, but 50 for other methods to maximize their performance. For each model, we performed hold-out analysis 10 times and reported the mean and standard deviation of clustering accuracies.

### 1.4 Derivation of Patterns Among Abeta+ MCI/Dementia Participants

To derive patterns observed specifically in Abeta+ participants, we selected out 145 CN Abeta- participants and 317 MCI/Dementia Abeta+ participants from the original discovery set to construct the new training set. We rerun the Smile-GANs model 30 times and used the same procedure introduced in section 4.6 in the main text to assign pattern probabilities to all Abeta+ participants including those who are not incorporated into the training set.

### 1.5 Prediction of Longitudinal MRI Progression Pathways

P2 and P3, described in the main text, were directly used for predicting longitudinal pattern stability or pattern progression pathways for P1 participants without special design of the model. Detailed experiments are introduced in the main text section 4.8. Participants with both P2 and P3 lower than optimal thresholds were predicted to remain in P1. For participants who have both P2 and P3 over thresholds, the true progression direction can be indicated by the pattern with higher probability. Only participants with P1 probability greater than 0.7 are included for analysis to avoid trivial prediction cases.

### 1.6 Comparison of P4 participants from Both Pathway

To compare P4 participants arriving along two different pathways, we utilized longitudinal data and focused on participants who had once reached P2 or P3 (P2>0.5 or P3>0.5). We selected and regrouped their later visits based on P4 probabilities, obtaining three subgroups with 0.5<P4<0.7, 0.7<P4<0.9, and P4>0.9 for each progression path respectively. AFNI 3dttest with GM tissue maps was then performed between CN and each subgroup.

## 2. SUPPLEMENTARY RESULTS

### 2.1 Smile-GAN Model Validation

#### 2.1.1 Synthetic Test

By assigning participants to the pattern type with the highest probability, the Smile-GAN model was able to cluster participants with 100% accuracy even with very severe confounding patterns. Moreover, Fig. 1a shows that mapping function $f$ is able to perfectly capture three simulated atrophy patterns while avoiding all non-disease-related patterns.

Figure 2b shows the change of WD and clustering error (i.e., 1 - clustering accuracy) during the training procedure. Generally, the two metrics are consistent in monitoring the training process. Therefore, WD could be used as a surrogate of clustering accuracy when the latter is not available in real applications. Note that inconsistencies at the beginning of training or oscillations exist. Such circumstances can be filtered out by the other two metrics, alteration quantity (AQ) and cluster loss. We propose to use WD, along with AQ and cluster loss, as metrics for monitoring the training process.

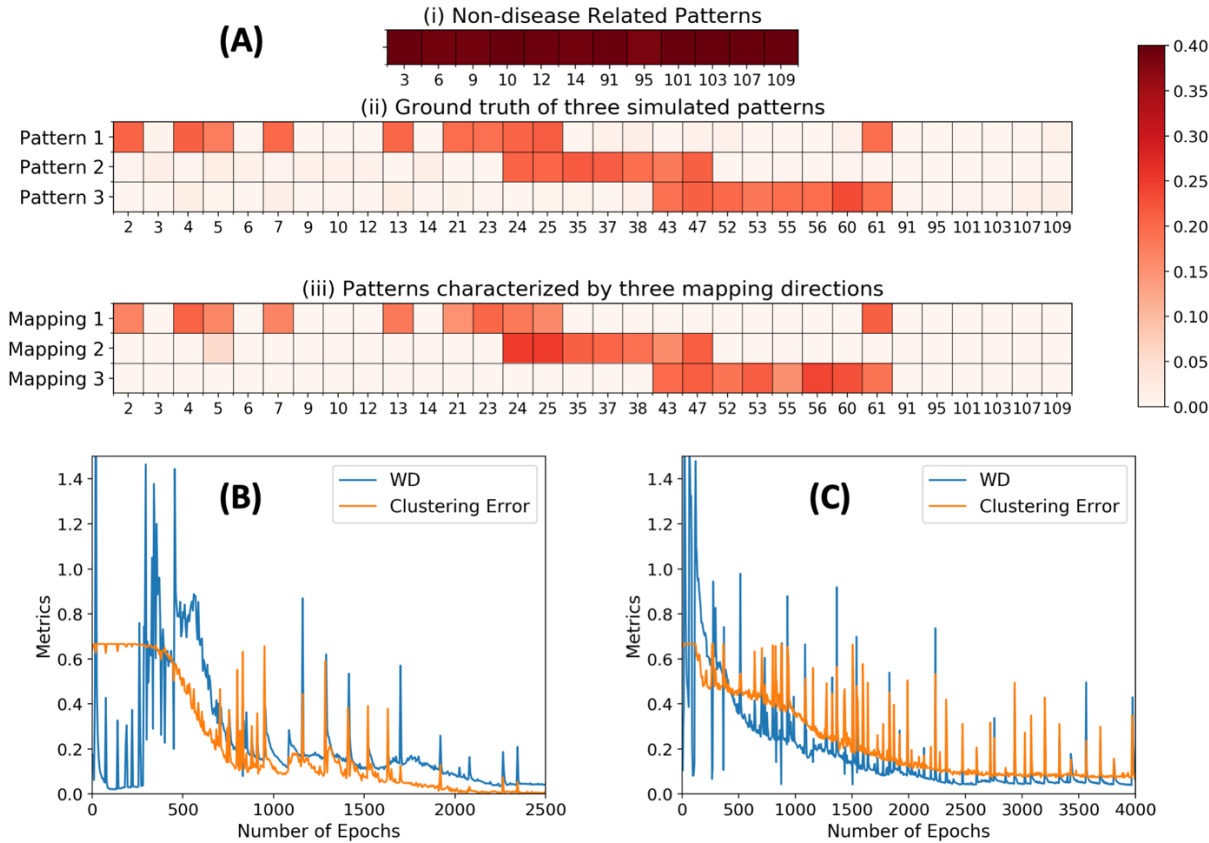

Fig. 1: Performance of Smile-GAN model on synthetic and semi-synthetic dataset. (a): The mapping function ($f$) of Smile-GAN discovers (iii) the ground truth of three simulated pure atrophy patterns (ii) while not confounded by very severe non-disease related patterns (i). Each mapping direction captures one type of pattern. Only the ROIs whose values decreasing over 5% are displayed. (b): Clustering training process monitoring with Wasserstein distance (WD) on synthetic test. Clustering error = 1- clustering accuracy. (c): Clustering training process monitoring with Wasserstein distance (WD) on semi-synthetic test. Clustering error = 1- clustering accuracy.

### 2.1.1 Synthesized Brain Atrophy Experiment

First, from Fig. 1c, we can find out that WD is still a good metric for training monitoring on data whose distribution is closer to real data. Also, As shown in Table 3, Smile-GAN outperforms all other semi-supervised or unsupervised clustering methods. It is shown to be more robust to tiny atrophy rates, overlaps of atrophy patterns and non-disease-related covariates.

Table 3: Clustering accuracy comparison between Smile-GAN and other methods

| Atrophy rate | Smile-GAN | HYDRA | CHIMERA | K-means | GMM |
|---|---|---|---|---|---|
| 0.3 | **0.999±0.001** | 0.972±0.005 | 0.582±0.013 | 0.450±0.036 | 0.449±0.037 |
| 0.1-0.3 | **0.958±0.009** | 0.877±0.037 | 0.381±0.006 | 0.288±0.007 | 0.294±0.004 |
| 0.1-0.2 | **0.825±0.017** | 0.550±0.045 | 0.371±0.004 | 0.254±0.040 | 0.272±0.002 |

### 2.2 Patterns Identified from Abeta+ MCI/Dementia Participants

As shown in Fig. 2a, four patterns derived from Abeta+ participants (referred as Abeta+ patterns) are closely correlated with the original four patterns. Therefore, the original 4-P system was able to express patterns observed in Abeta+ participants, since four Abeta+ patterns are only shifted forward to a more severe stage from the original four patterns along two pathways (Fig. 2b).

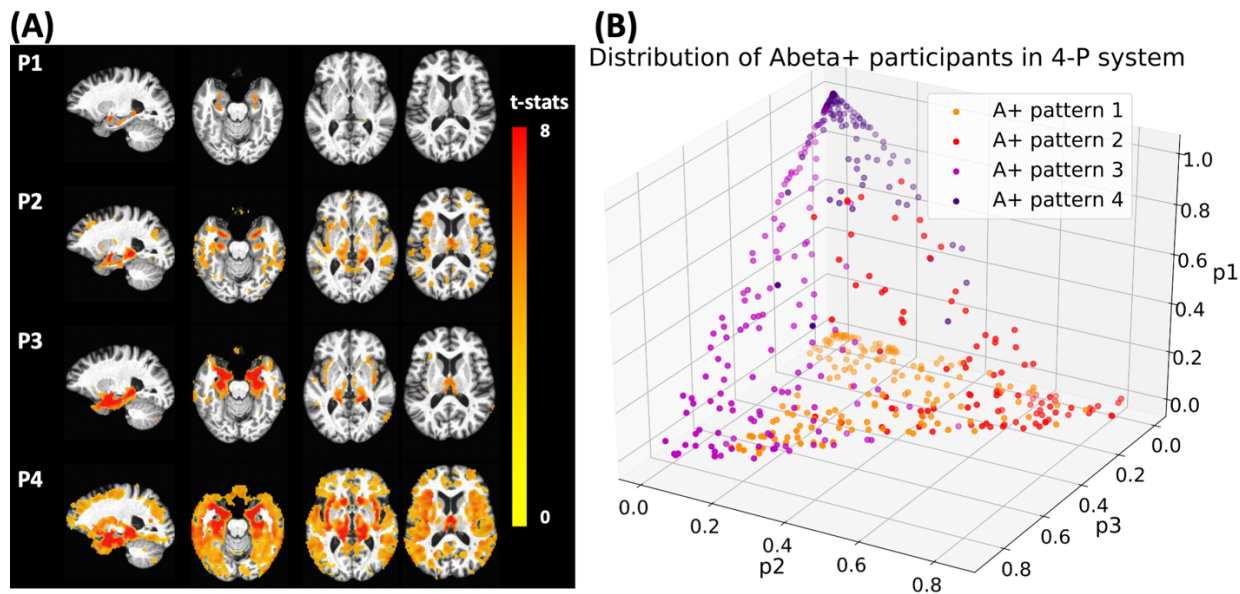

Fig. 2: Four Abeta+ patterns and their connection with original four patterns. (a) Voxel-wise statistical comparison between CN and Abeta+ participants predominantly belonging to the four Abeta+ patterns. FDR correction for multiple comparisons with p-value threshold of 0.05 was applied. (b) Visualization of Abeta+ participants' expression of original four patterns in the same 3d system. Colors of each dot are determined by the predominant Abeta+ pattern.

## 2.3 Demographics, Clinical biomarkers and Cognitive scores

Clinical characteristics of different groups of participants are revealed in Table 4,5,6.

| Table 4 (1) | P1 | P2 | P3 | P4 |
|---|---|---|---|---|
| TTau | 294.2 (212.0-398.52) (104) | 297.5 (204.9-406.9) (117) | 341.05 (267.85-446.7) (144) | 306.35 (242.58-404.02) (232) |
| PTau | 27.52 (20.03-40.19) (104) | 30.48 (20.08-40.9) (117) | 34.87 (26.11-46.43) (144) | 30.3 (23.94-39.36) (232) |
| Age | 72.08 (66.74-77.82) (104) | 73.68 (66.48-77.66) (117) | 73.86 (70.07-77.83) (144) | 75.86 (71.15-79.62) (232) |
| WML | 34.1 (25.93-49.55) (67) | 46.32 (30.18-59.15) (78) | 45.13 (33.63-57.24) (75) | 50.62 (39.12-66.06) (127) |
| ADNI-EF | 0.23 (-0.29-0.97) (86) | -0.29 (-0.82-0.32) (94) | -0.1 (-0.77-0.63) (121) | -0.71 (-1.52--0.11) (190) |
| ADNI-MEM | 0.21 (-0.22-0.63) (86) | -0.13 (-0.5-0.41) (94) | -0.39 (-0.75-0.07) (121) | -0.8 (-1.17--0.3) (190) |
| ADNI-LAN | 0.26 (-0.15-0.87) (86) | -0.04 (-0.47-0.4) (94) | -0.11 (-0.7-0.44) (121) | -0.62 (-1.34--0.08) (190) |
| Hippocampal volume | 0.61 (0.58-0.64) (104) | 0.60 (0.57-0.63) (117) | 0.54 (0.51-0.56) (144) | 0.52 (0.48-0.56) (232) |
| Gender (% of male) | 0.61 | 0.62 | 0.5 | 0.63 |
| % of Tau+ | 60.61% | 64.35% | 78.42% | 73.42% |
| ApoE ε4 Carriers | 67.31% | 52.14% | 78.47% | 69.83% |
| MCI | 92 | 89 | 97 | 102 |
| Dementia | 12 | 28 | 47 | 130 |

| Table 4 (2) | P1vsP4 | P1vsP2 | P1vsP3 | P2vsP3 | P2vsP4 | P3vsP4 |
|---|---|---|---|---|---|---|
| TTau | 0.542 | 0.994 | **0.002** | **0.002** | 0.534 | **<0.001** |
| PTau | 0.684 | 0.534 | **0.001** | **0.01** | 0.672 | **<0.001** |
| Age | **<0.001** | 0.547 | **0.01** | 0.077 | **<0.001** | **0.028** |
| WML | **<0.001** | **0.031** | 0.039 | 0.863 | **0.047** | **0.028** |
| ADNI-EF | **<0.001** | **<0.001** | **0.009** | **0.038** | **<0.001** | **<0.001** |
| ADNI-MEM | **<0.001** | **0.003** | **<0.001** | **0.003** | **<0.001** | **<0.001** |
| ADNI-LAN | **<0.001** | **0.002** | **<0.001** | 0.166 | **<0.001** | **<0.001** |
| Hippocampal volume | **<0.001** | 0.162 | **<0.001** | **<0.001** | **<0.001** | **0.025** |
| Gender (% of male) | 0.83 | 0.89 | 0.128 | 0.06 | 0.922 | **0.023** |
| % of Tau+ | **0.03** | 0.673 | **0.005** | **0.019** | 0.109 | 0.344 |
| ApoE ε4 Carriers | 0.738 | **0.031** | 0.068 | **<0.001** | **0.002** | 0.086 |

Table 4 (1) and (2): Demographics, clinical biomarkers and cognitive scores of participants who are diagnosed as MCI or Dementia and have positive Abeta status at baseline. Hippocampal volume is normalized with respect to 0.01*total brain volume. Median (first quartile – third quartile) (number of participants) are reported. For categorial variables, chi-squared test was used to identify differences between subgroups. For other quantitative variables, an ANOVA analysis was performed for comparison.

|  | CN (1) | MCI/Dementia (2) | 1 vs 2 |
|---|---|---|---|
| TTau | 338.3 (255.55-389.55) (19) | 346.45 (280.9-446.7) (160) | 0.329 |
| PTau | 33.43 (25.78-41.19) (19) | 35.46 (26.97-46.8) (160) | 0.481 |
| Age | 76.78 (73.73-80.23) (19) | 73.86 (69.71-78.07) (160) | **0.021** |
| Education year | 18.0 (16.5-19.0) (19) | 16.0 (13.75-18.0) (160) | **0.001** |
| Hippocampal volume | 0.0055 (0.53-0.57) (19) | 0.0054 (0.51-0.56) (160) | **0.018** |
| P3 Probability | 0.53 (0.48-0.61) (19) | 0.61 (0.49-0.71) (160) | **0.048** |

Table 5: Demographics, clinical biomarkers of participants who are dominated by P3 and have positive Abeta status at baseline. Hippocampal volume is normalized with respect to 0.01*total brain volume. Median (first quartile – third quartile) (number of participants) are reported and all comparisons were performed using an ANOVA test.

|  | A-/T- (1) | A+/T- (2) | A+/T+ (3) | 1 vs 2 | 1 vs 3 | 2 vs 3 |
|---|---|---|---|---|---|---|
| TTau | 199.3 (165.95-229.2) (87) | 205.8 (164.65-217.5) (35) | 378.3 (310.18-483.15) (64) | 0.625 | **<0.001** | **<0.001** |
| PTau | 17.2 (14.67-19.64) (87) | 19.07 (15.68-20.6) (35) | 39.08 (31.06-48.14) (64) | 0.11 | **<0.001** | **<0.001** |
| Age | 67.95 (62.83-73.66) (87) | 71.87 (65.33-78.07) (35) | 72.6 (67.48-77.82) (64) | **0.045** | **0.008** | 0.904 |
| WML | 32.0 (22.37-42.05) (67) | 31.3 (25.1-51.92) (27) | 38.2 (26.65-49.06) (34) | 0.1 | **0.023** | 0.753 |
| ADNI-EF | 0.71 (0.19-1.37) (71) | 0.65 (-0.01-1.08) (28) | 0.07 (-0.39-0.54) (53) | 0.199 | **<0.001** | **0.03** |
| ADNI-MEM | 0.56 (0.18-1.22) (71) | 0.38 (0.03-0.74) (28) | 0.02 (-0.34-0.52) (53) | 0.091 | **<0.001** | **0.028** |
| ADNI-LAN | 0.56 (0.07-1.27) (71) | 0.46 (0.06-0.87) (28) | 0.18 (-0.21-0.6) (53) | 0.36 | **0.006** | 0.129 |
| Hippocampal volume | 0.63 (0.59-0.66)*10 (87) | 0.61 (0.59-0.65) (35) | 0.6 (0.56-0.63) (64) | 0.716 | **<0.001** | **0.011** |
| P1 Probability | 0.84 (0.66-0.95) (87) | 0.87 (0.8-0.97) (35) | 0.74 (0.57-0.91) (64) | **0.012** | **0.028** | **<0.001** |
| Gender (% of male) | 48.28% | 71.43% | 57.81% | 0.034 | 0.32 | 0.262 |
| ApoE ε4 Carriers | 18.39% | 51.43% | 75.00% | **<0.001** | **<0.001** | **0.031** |
| MCI | 84 | 34 | 54 |  |  |  |
| Dementia | 3 | 1 | 10 |  |  |  |

Table 6: Demographics, clinical biomarkers and cognitive scores of participants who are dominated by P1 and diagnosed as MCI or Dementia at baseline. Hippocampal volume is normalized with respect to 0.01*total brain volume. Median (first quartile – third quartile) (number of participants) are reported. For categorial variables, chi-squared test

was used to identify differences between subgroups. For other quantitative variables, an ANOVA analysis was performed for comparison. (A: Abeta; T: pTau)

## 2.4 Prediction Results of MRI Progression Pathways

As also mentioned in the main text, prediction performance of longitudinal pattern progression worsens after the fifth year and optimal thresholds to predict progression along either pathway decrease with time. Moreover, among participants correctly classified as progression in either direction, there is a fraction of them who have both P2 and P3 over thresholds as shown in Table 7 and, thus, their true progression pathway cannot be clearly revealed by optimal thresholds. For these participants, the true progression direction can be simply indicated by the pattern with higher probability with high accuracies.

|  | 2y | 3y | 4y | 5y | 6y | 7y | 8y |
|---|---|---|---|---|---|---|---|
| **P1-P2** | | | | | | | |
| **AUC** | 0.857 | 0.882 | 0.851 | 0.848 | 0.843 | 0.824 | 0.807 |
| **Threshold** | 0.141 | 0.128 | 0.106 | 0.081 | 0.078 | 0.078 | 0.065 |
| **Accuracy** | 0.868 | 0.849 | 0.791 | 0.751 | 0.753 | 0.745 | 0.701 |
| **P1-P3** | | | | | | | |
| **AUC** | 0.904 | 0.974 | 0.900 | 0.907 | 0.855 | 0.845 | 0.839 |
| **Threshold** | 0.078 | 0.080 | 0.058 | 0.044 | 0.037 | 0.037 | 0.034 |
| **Accuracy** | 0.903 | 0.916 | 0.852 | 0.830 | 0.779 | 0.766 | 0.749 |
| **Participants with both P2 and P3 over thresholds** | | | | | | | |
| **Quantity** | 14.3% | 9.7% | 21.4% | 27.1% | 27.6% | 29.3% | 34.5% |
| **Accuracy** | 100% | 100% | 100% | 93.8% | 95.2% | 95.8% | 90% |

Table 7: Prediction results of longitudinal progression pathways. AUC and optimal thresholds are reported as described in Section 1.5; Accuracies for P1-P2 and P1-P3 predictions at different time points were derived with optimal thresholds. Quantity shows the fraction of participants who are correctly predicted as progression in either direction but actually have both P2 and P3 over corresponding thresholds. Basing prediction on the higher probability between P2 and P3 when both are above threshold results in accuracy as shown in the last row.

## 2.5 Comparison of P4 Participants from Both Pathway

As P4 probabilities approach 1, both pathways converge to one common pattern type with atrophy in almost consistent regions, though there are still differences in the severity of atrophy in some regions, like the medial temporal lobe.

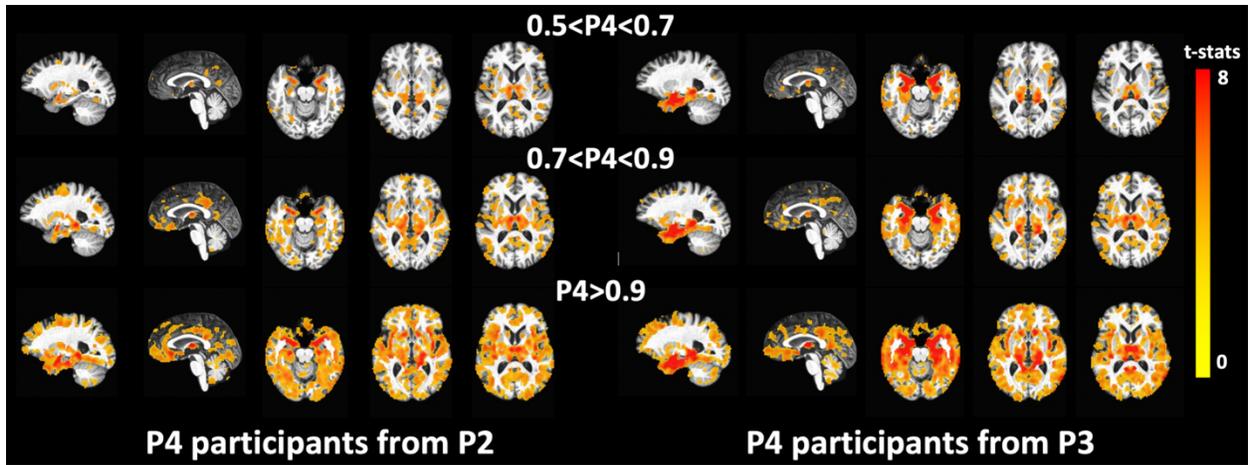
Fig. 3: Atrophy development of P4 participants from both pathways. Voxel-wise statistical comparison between CN and 6 subgroups obtained through procedures mentioned in method section 1.6. FDR correction for multiple comparisons with p-value threshold of 0.05 was applied.

# Supplementary Reference: